\documentclass[11pt]{article}

\usepackage{graphicx} 
\usepackage{subcaption} 
\usepackage[utf8]{inputenc}
\usepackage[T1]{fontenc}
\usepackage{microtype}
\usepackage[a4paper]{geometry}
\usepackage{amsmath,amsfonts,amsthm,amssymb,mathtools}
\usepackage{tikz}
\usetikzlibrary{positioning, arrows, patterns}
\usepackage{float}
\usepackage{algorithm}
\usepackage{algpseudocode}
\usepackage[hidelinks]{hyperref}
\usepackage{enumerate}
\usepackage{url}
\usepackage{bm}
\usepackage[affil-it]{authblk}

\usepackage{orcidlink}
\usepackage{booktabs, multirow}
\usepackage{rotating}
\usepackage{csquotes}
\MakeOuterQuote{"}
\usepackage{comment}

\newcommand{\R}{\mathbb{R}}
\newcommand{\N}{\mathbb{N}}
\renewcommand{\H}{\mathbb{H}}
\newcommand{\EE}{\mathbb{E}}
\newcommand{\LL}{\mathbb{L}}
\newcommand{\X}{\bm{X}}
\newcommand{\Y}{\bm{Y}}
\newcommand{\Z}{\bm{Z}}
\newcommand{\z}{\bm{z}}
\newcommand{\x}{\bm{x}}
\newcommand{\V}{\bm{V}}
\newcommand{\hV}{\widehat{\bm{V}}}
\newcommand{\W}{\bm{W}}
\newcommand{\w}{\bm{w}}
\renewcommand{\v}{\bm{v}}
\renewcommand{\u}{\bm{u}}
\renewcommand{\S}{\bm{S}}
\newcommand{\U}{\bm{U}}

\newcommand{\de}{\stackrel{\mathrm{def}}{=}}
\newcommand{\lp}{\left(}
\newcommand{\rp}{\right)}
\newcommand{\simplex}{\Delta_{d-1}}
\renewcommand{\to}{\rightarrow}
\renewcommand{\P}{\mathsf{P}}
\newcommand{\Q}{\mathsf{Q}}
\newcommand{\E}{\mathsf{E}}
\newcommand{\wc}{\stackrel{\mathcal{L}}{\to}}
\newcommand{\sigvec}{\boldsymbol{\sigma}}
\newcommand{\xivec}{\boldsymbol{\xi}}
\newcommand{\gamvec}{\xivec}
\newcommand{\0}{\bm{0}}
\newcommand{\1}{\bm{1}}
\newcommand{\diff}{\mathrm{d}}
\newcommand{\I}{\mathbb{I}}
\newcommand{\Lip}{\mathrm{Lip}}
\DeclareMathOperator{\clr}{clr}
\newcommand{\e}{\bm{e}}
\newcommand{\gen}{\mathcal{G}}
\newcommand{\test}{\mathcal{T}}
\newcommand{\bTheta}{\bm{\Theta}}
\newcommand{\interior}[1]{%
  {\kern0pt#1}^{\mathrm{o}}%
}
\newcommand{\osimplex}{\interior{\Delta}_{d-1}}
\newcommand{\MGP}{\bm{H}}
\newcommand{\approxlaw}{\stackrel{\mathcal{L}}{\approx}}
\newcommand{\eqlaw}{\stackrel{\mathcal{L}}{=}}

\newcommand{\tPhi}{\widetilde{\Phi}}
\newcommand{\hPhi}{\widehat{\Phi}}

\theoremstyle{plain}

\newtheorem{proposition}{Proposition}

\newtheorem{assumption}{Assumption}

\theoremstyle{remark}
\newtheorem{remark}{Remark}

\usepackage{color}
\definecolor{darkteal}{rgb}{0,0.35,0.35}

\definecolor{changecol}{rgb}{0,0,0}
\newcommand{\chg}[1]{\textcolor{changecol}{#1}}
\newcommand{\rev}[1]{\bgroup \color{black} #1 \egroup}

\newcommand{\keywords}[1]{%
  \vspace{1em}
  \noindent\textbf{\small Keywords:}~{\small #1}
}

\title{\rev{Simulation of Multivariate Extremes: \\ a Wasserstein--Aitchison GAN approach}}

\author[1,4]{Stéphane Lhaut \orcidlink{0000-0003-2449-6218} \thanks{\href{mailto:stephane.lhaut@ensae.fr}{stephane.lhaut@ensae.fr}, corresponding author}}
\author[2]{Holger Rootzén \orcidlink{0000-0001-8869-7989} \thanks{\href{mailto:hrootzen@chalmers.se}{hrootzen@chalmers.se}}}
\author[1,3]{Johan Segers \orcidlink{0000-0002-0444-689X} \thanks{\href{mailto:jjjsegers@kuleuven.be}{jjjsegers@kuleuven.be}}}

\affil[1]{Institute of Statistics, Biostatistics and Actuarial Sciences, LIDAM, UCLouvain, Voie du Roman Pays, 20, 1348 Louvain-La-Neuve, Belgium}
\affil[2]{Department of Mathematical Sciences, Chalmers University of Technology and University of Gothenburg, Gothenburg, Sweden}
\affil[3]{Department of Mathematics, KU Leuven, Celestijnenlaan 200B, 3001 Heverlee, Belgium}
\affil[4]{CREST, ENSAE Paris, 5 avenue Henry Le Chatelier 91120 Palaiseau, France}

\date{\today}

\begin{document}

\maketitle

\begin{abstract}
\rev{Economically responsible mitigation of multivariate extreme risks—such as extreme rainfall over large areas, large simultaneous variations in many stock prices, or widespread breakdowns in transportation systems—requires assessing the resilience of the systems under plausible stress scenarios.  This paper uses Extreme Value Theory (EVT) to develop a new approach to simulating such multivariate extreme events. 
Specifically, we assume that after transformation to a standard scale the distribution of the random phenomenon of interest is multivariate regular varying and use this to provide a sampling procedure for extremes on the original scale.
Our procedure combines a Wasserstein–Aitchison Generative Adversarial Network (WA-GAN) to simulate the tail dependence structure on the standard scale with joint modeling of the univariate marginal tails on the original scale. 
The WA-GAN procedure relies on the angular measure---encoding the distribution on the unit simplex of the angles of extreme observations---after transformation to Aitchison coordinates, which allows the Wasserstein-GAN algorithm to be run in a linear space. Our method is applied both to simulated data under various tail dependence scenarios and to a financial data set from the Kenneth French Data Library. The proposed algorithm demonstrates strong performance compared to existing alternatives in the literature, both in capturing tail dependence structures and in generating accurate new extreme observations.}
\end{abstract}


\section{Introduction}
\label{sec:intro}

\paragraph*{Multivariate EVT.}
\rev{
	Dependence between extreme events is often of crucial importance. Extreme rainfall is even more dangerous if it occurs simultaneously at many spatial locations; costs associated with breakdowns in transportation systems are more harmful if they occur in many parts of the system; and extreme fluctuations in financial markets can cause much more damage if they occur across many sectors of the economy. Over the last decades, statistical research on dependent, multivariate extremes has grown rapidly and now includes many useful, mostly parametric, models and methods. Often, the computational challenges associated with estimating the parameters of these models restrict their use to relatively low-dimensional settings. When spatial information can be incorporated into the model, dimensions may be higher, but this is limited to specific applications, mainly climatic ones. For background on EVT, we refer to standard monographs such as \cite{beirlant2004, deHaanFerreira2006, Resnick2007} or more recent ones such as \cite{mikosch2024evt, resnick2024hidden}.
}

\paragraph*{GenAI.}

\rev{
	Generative Artificial Intelligence (GenAI) methods based on artificial neural networks now provide completely new opportunities to address challenges in high dimensions. The GenAI approach differs from the standard statistical modeling perspective in that the output of these methods is a sampling algorithm rather than a set of parameters from a given parametric model, from which sampling can be performed in a second stage. Here, no explicit assumption on the data distribution is made, even though certain assumptions may help in designing the neural network architectures. One of the most popular such algorithms is the Generative Adversarial Network (GAN) \cite{goodfellow2014} or its variant, the Wasserstein-GAN (WGAN) \cite{arjovsky2017}. See~\cite[Chapter~17]{bishop2024deep} for a pedagogical introduction to GANs and their variations. Basically, the idea is to train two neural networks simultaneously: one aims to generate new realizations of the random phenomenon of interest (the generator) by transforming samples from a known (latent) distribution, while the other attempts to discriminate (the discriminator) between the true and generated realizations. The difference between GAN and WGAN lies in the distance used to compare samples, which is related to the Jensen--Shannon divergence in the case of GAN and to the Wasserstein distance in the case of WGAN. The latter aims to avoid the well-known ``mode collapse'' issues associated with GANs and is therefore often used as the default choice in such approaches \cite{arjovsky2017}. Initially, GAN algorithms were proposed for synthetic image generation, but they have since been applied to many types of data, in particular tabular data, which are of special interest to statisticians~\cite{annoye2025}. Some theoretical properties of the WGAN algorithm can be found in \cite{biau2021wgans}. Many other sampling algorithms exist in the literature, such as Normalizing Flows (NF), Variational Autoencoders (VAE), or Diffusion Models (DM), each with its own merits and drawbacks. As our method is based on the WGAN algorithm, we do not discuss these alternative approaches in detail here, but the interested reader may find an introduction to them in \cite[Chapters~18, 19, 20]{bishop2024deep}.
}

\paragraph*{Bridging EVT and GenAI.}

\rev{
	Using GenAI approaches to simulate new multivariate extreme scenarios is challenging for multiple reasons. First, it is well known from EVT that the tail dependence structure of a random vector may differ drastically from its central dependence. As such, a straightforward application of a sampling algorithm to the entire dataset has little chance of accurately capturing the tail behavior of the underlying data-generating process. Second, as pointed out in \cite{Girard-et-al}, models based on transformations of light-tailed noise, such as GANs, fail to capture the tail behavior of heavy-tailed distributions, and these algorithms must therefore be adapted. Lastly, GenAI approaches typically require large amounts of data, whereas extreme events are, by definition, rare. Consequently, the algorithms need to be adapted in order to deal with the limited amount of data relevant for tail problems.
	
	The use of EVT methods within GenAI frameworks for simulating accurate new extremes has attracted considerable attention recently. Adaptations of GAN algorithms have been considered in \cite{Allouche-et-al-EV-GAN, Allouche-et-al-excessGAN, exGAN-bathia-et-al2021, Huster2021ParetoGAN}. These approaches all rely on univariate EVT methods applied to each margin, but they do not incorporate tail dependence structures arising from multivariate EVT. An inherently multivariate approach is proposed in \cite{Boulagiem-et-al}, which applies univariate EVT to the margins and models the dependence of the full dataset after a copula transformation. No multivariate EVT is used there, as is also the case in \cite{Girard-et-al}, whose Heavy-Tailed GAN (HTGAN) method aims to model dependence after transformation to heavy-tailed margins using a heavy-tailed latent distribution for the generator. GAN approaches that explicitly make use of multivariate EVT can be found in the Generalized Pareto GAN (GPGAN) algorithm \cite{Li-et-al}, which targets multivariate threshold exceedances based on multivariate Generalized Pareto (MGP) distributions \cite{rootzen2006}, or in the $d$-max-decreasing Neural Networks (dMNN) approach \cite{dMNNs-hasan-et-al}, which relies on multivariate block maxima and the Pickands dependence function, see \cite[Chapter~8]{beirlant2004} or \cite[Chapter~6]{deHaanFerreira2006}.

    Other approaches to sampling multivariate extremes can be found in the literature, some of which are very recent. A VAE-based method is proposed in \cite{VAE-Lafon-et-al}. Normalizing flow approaches in the context of ``geometric extremes'' (a framework different from multivariate regular variation) are presented in \cite{hu2025gpdflowgenerativemultivariatethreshold} and \cite{demonte2025generative}. Issues related to generating heavy-tailed outputs using normalizing flows are also discussed in \cite{Hikle-et-al-TFF}. We will not compare our method to these approaches, as our focus is on GAN algorithms and their variants.
	A broader comparison of many existing generative methods for extremes is provided in \cite{wessel2025comparisongenerativedeeplearning}.

    
    \paragraph*{WA-GAN.}
	The method we propose, Wasserstein--Aitchison GAN (WA-GAN), also relies on multivariate EVT and takes advantage of the well-known polar decomposition of a multivariate regularly varying random vector into an independent radial component and an angular component, which we represent by a point on the unit simplex.
    Since we work on a standard scale, no assumptions on the margins are required apart from continuity.
    The generative component of our method operates on the angular values. The advantage of this approach is that, at the angular scale, the data are bounded, and we can rely on Aitchison coordinates, a tool from compositional data analysis, to simulate values on the unit simplex by applying a WGAN on a linear space.
    The generative algorithm is then combined with standard marginal EVT modeling to yield a generative algorithm for multivariate threshold exceedances within the MGP framework. 
    
    We compare our approach to GPGAN, which follows a similar strategy but models both the marginal tails and the tail dependence structure simultaneously, and to HTGAN, which uses the same standardization as our method but then applies a standard GAN with heavy-tailed input to model dependence, without exploiting the specific structure of tail dependence.
	
}

\paragraph*{Outline.}

\rev{
	Section~\ref{sec:background} presents the background required to introduce our WA-GAN algorithm, together with a theoretical result (Proposition~\ref{prop:MGPapprox}) that provides the tools needed to transform angular data on the simplex into new multivariate extreme scenarios. Section~\ref{sec:methodo} describes the details of WA-GAN and presents the corresponding algorithms.
    Section~\ref{sec:architectures} describes  how performance of algorithms was evaluated  and the architecture used for WA-GAN. In Section~\ref{sec:simus}, WA-GAN is compared to HTGAN \cite{Girard-et-al} and GPGAN \cite{Li-et-al} using simulated data with logistic and H\"usler--Reiss dependence structures \cite[Section 9.2.2]{beirlant2004} in dimensions $d \in \{10, 20, 50\}$. Section~\ref{sec:simus:gaussian} discusses how the use of non-Gaussian latent distributions for the generator can help improve results when asymptotic independence is suspected. In Section~\ref{sec:real}, we apply WA-GAN to financial data consisting of daily returns for $d = 30$ industry portfolios. Section~\ref{sec:conclusions} summarizes the results.
}

\section{Background}
\label{sec:background}

\subsection{Regular variation and extreme value analysis}
\label{sec:regular-variatio}

\paragraph{Multivariate regular variation and angular measure.}
Let $\X = (X_1, \ldots, X_d)$ be the random vector of interest with values in $\R^d$. Based on a random sample of the (unknown) distribution of $\X$, we wish to generate new samples with the same tail behavior as $\X$, even at levels beyond those encountered in the sample. Such extrapolation is possible only if we are willing to make some regularity assumptions.
Our core assumptions to obtain accurate modeling of the tail of $\X$ are founded in the theory of multivariate regular variation~\cite{resnick1987}.
There are two approaches to use this hypothesis.
\begin{enumerate}[1.]
    \item Either we postulate it on the random vector $\X$ itself. This implies that all univariate tail indices are same.
    \item Or we postulate it on a transformed vector $\V = v(\X)$, where $v$ is a coordinatewise transformation of $\X$ so that the $d$ components of $\V$ all have the same distribution. This implies that the multivariate assumption only depends on the dependence structure of $\X$ and not on the margins, which can be assessed separately.
\end{enumerate}
Here we choose the second approach and transform to unit-Pareto margins using the transformation
\[
    v : \x = \lp x_j \rp_{j=1}^d \mapsto v(\x) \de \lp \frac{1}{1-F_j(x_j)} \rp_{j=1}^d,
\]
where $F_j$ denotes the distribution function of $X_j$.  We assume throughout that these marginal distribution functions are  continuous. It is straightforward to see that after this transformation the margins of $\V = v(\X)$ have unit-Pareto distributions, $\P(V_j > y) = 1/y$ for $y \ge 1$. Our first assumption describes the asymptotic dependence structure of $V$.

\begin{assumption}[Multivariate regular variation]
\label{ass:multRV}
There exists a non-zero Borel measure $\nu$ on $\EE \de [0,\infty)^d \setminus \{\0\}$ which is finite on Borel sets bounded away from $\0$ and such that
\[
    \lim_{t \to \infty} t \ \P \lp t^{-1} \V \in B \rp = \nu(B)
\]
for all Borel sets $B$ of $\EE$ bounded away from $\0$ with $\nu(\partial B) = 0$, with $\partial B$ the topological boundary of $B$.
\end{assumption}

Intuitively, the measure $\nu$ helps in describing the tail behavior of $\V$ as for large $t>0$, we have $\P(\V \in tB) \approx \nu(B)/t$ where $tB \de \{t \x: \x \in B\}$. The measure $\nu$ is often referred to as the \emph{exponent measure} of $\V$, because of its appearance in the exponent of the expression of the multivariate extreme value distribution to which the law of $\V$ is attracted~\cite[Definition~6.1.7]{deHaanFerreira2006}.

Our procedure below will benefit from an equivalent formulation of Assumption~\ref{ass:multRV} in polar coordinates with respect to the $L_1$-norm $|\x|_1 \de \sum_{j=1}^d |x_j|$ for $\x \in \R^d$. More concretely, it can be shown~\cite[Theorem~6.1]{Resnick2007} that due to a convenient homogeneity property of $\nu$~\cite[Theorem~6.1.9]{deHaanFerreira2006}, if we put
\begin{equation}
\label{eq:polar}
    R \de |\V|_1  \qquad \text{and} \qquad \W \de R^{-1} \V,
\end{equation}
where $\W$ takes values in the unit simplex $\simplex \de \{\x \in [0, 1]^d : |\x|_1 = 1 \}$, there exists a probability measure $\Phi$ on $\simplex$ such that
\begin{equation}
\label{eq:RVpolar}
    \lim_{t \to \infty} \P \lp \W \in A \mid R \geq t \rp = \Phi(A),
\end{equation}
for any Borel set $A \subseteq \simplex$ such that $\Phi(\partial A) = 0$.

The measure $\Phi$ is often referred to as the \emph{angular (probability) measure} of $\V$ as it describes how its ``angular'' component $\W$ behaves when its radial part $R$ becomes large. The set of possible probability measures that can be obtained in such a way forms a nonparametric class as the only constraint they should satisfy---due to the unit-Pareto margins of $\V$---are
\begin{equation}
\label{eq:marginalPhi}
    \int_{\simplex} w_j \, \diff\Phi(\w) = \frac{1}{d}, \qquad j \in \{1,\ldots,d\},
\end{equation}
wich in turn implies
\begin{equation}
\label{eq:Rtail}
   \P(R>t) \sim \frac{d}{t},  \qquad \text{as } t \to \infty.
\end{equation}

It is possible for $\Phi$ to have positive mass on one of the faces of the simplex $\simplex$, that is, sets for which at least one of the coordinates is zero. In terms of extremes, it corresponds to scenarios where some components of $\X$ are large while others are small. These are called \emph{extreme directions} and have to be treated by specific methods which our methodology is not able to handle, see, e.g., \cite{mourahib2024directions}. We exclude this situation here.

\begin{assumption}
\label{ass:phi_interior}
The measure $\Phi$ in~\eqref{eq:RVpolar} is concentrated on the open simplex $\osimplex \de \{\x \in (0,1)^d : |\x|_1 = 1\}$, that is, if $\bTheta \sim \Phi$, then
\[
    \P \lp \bTheta \in \osimplex \rp = 1.
\]
\end{assumption}

\paragraph{Univariate generalized Pareto distributions.}
Assumption~\ref{ass:multRV} is not sufficient by itself to model the tail behavior of $\X$ as it only concerns the dependence structure. Our second main assumption concerns the tails of the margins $X_j$ and allows for heterogeneous tails. Here and below, $u_{*j}$ denotes the (possibly infinite) upper endpoint of $X_j$.

\begin{assumption}[Generalized Pareto limit of marginal tails]
\label{ass:margins_rv}
For every $j \in \{1,\ldots,d\}$, there exist $\xi_j \in \R$ and a function $\sigma_j(\cdot) : (-\infty,u_{*j}) \to (0,\infty)$ such that for all $y > 0$,
\[
    \lim_{u \nearrow u_{*j}} \P \lp \frac{X_j - u}{\sigma_j(u)} > y \, \big| \, X_j > u \rp
    = \lp 1 + \xi_j y \rp^{-1/\xi_j}_+,
\]
where $a_+ \de \max(a,0)$ for $a \in \R$; if $\xi_j = 0$, the limit is to be understood as $\exp(-y)$.
\end{assumption}


By the Pickands--Balkema--de Haan theorem \cite{balkema1974residual,pickands1975statistical}, Assumption~\ref{ass:margins_rv} is equivalent to the assumption that $X_j$ is in the maximum domain of attraction of a GEV distribution.
The limit in Assumption~\ref{ass:margins_rv} holds uniformly in $y > 0$. 
The assumption justifies the following approximation for $y>0$ such that $1 + \xi_j y / \sigma_j > 0$:
\begin{equation}
\label{eq:GPapprox}
    \P \lp X_j - u > y \, \big| \, X_j > u \rp
    \approx \lp 1 + \frac{\xi_j y}{\sigma_j} \rp^{-1/\xi_j},
\end{equation}
where $\sigma_j > 0$ is a scale parameter that accounts for the unknown function $\sigma_j(\cdot)$. Estimation of the bivariate parameter $(\sigma_j,\xi_j)$ is typically performed by the method of maximum likelihood. This is the well-known \emph{Peaks-Over-Threshold (PoT)} approach for modeling the tails of real-valued data, see for instance \cite{smith1987estimating} and \cite[Section~5.3]{beirlant2004}.
A random variable with distribution function
\[
    H_{\sigma,\xi}(y) \de 1 - \lp 1 + \frac{\xi y}{\sigma} \rp^{-1/\xi}_+, \qquad y > 0,
\]
for some $(\sigma, \xi) \in (0,\infty) \times \R$ is said to have a \emph{generalized Pareto (GP) distribution}. 

 high-risk scenarios is the case where

\begin{remark}[Equivalent formulations of Assumption~\ref{ass:margins_rv}]
\label{rem:equiv_margins}
As already explained, Assumption~\ref{ass:margins_rv} is equivalent to the hypothesis that $X_j$ is in the maximum domain of attraction of a GEV distribution. As a consequence, $X_j$ satisfies all the equivalent formulations of this condition as presented, e.g., in Theorem~1.1.6 of~\cite{deHaanFerreira2006}. As some of them will be needed below, we recall them here. Let $b_j(t) \de F_j^{-1}(1-1/t) = (1/(1-F_j))^{-1}(t)$, for $t > 1$, be the tail quantile function of $F_j$. Then, the following statements are equivalent to Assumption~\ref{ass:margins_rv}.
\begin{enumerate}[(i)]
    \item There exist $\xi_j \in \R$ and a scaling function $a_j(\cdot) : (1,\infty) \to (0,\infty)$ such that
    \[
    \lim_{t \to \infty}
    \P \left( \frac{X_j - b_j(t)}{a_j(t)} > y \mid X_j > b_j(t) \right)
    = \lp 1 + \xi_j y \rp^{-1/\xi_j}_+, \qquad y > 0.
    \]
    \item There exist $\xi_j \in \R$ and a scaling function $a_j(\cdot) : (1,\infty) \to (0,\infty)$ such that
    \[
        \lim_{t\to\infty} \frac{b_j(ty)-b_j(t)}{a_j(t)} = \frac{y^{\xi_j}-1}{\xi_j},
        \qquad y > 0,
    \]
    where the right-hand side is to be interpreted as $\log y$ if $\xi_j = 0$. The convergence also holds locally uniformly in $y \in (0, \infty)$, see Section~1.2 in~\cite{deHaanFerreira2006}.
\end{enumerate}
The scalar $\xi_j \in \R$ in (i) and (ii) is equal to the one in Assumption~\ref{ass:margins_rv}, while the scaling function $a_j(\cdot)$ in (i) and (ii) is related to $\sigma_j(\cdot)$ in Assumption~\ref{ass:margins_rv} via $\sigma_j(u) = a_j(1/(1-F_j(u)))$ for $u<u_{*j}$.
\end{remark}

\paragraph{Multivariate generalized Pareto distributions.}
What do Assumptions~\ref{ass:multRV}--\ref{ass:margins_rv}
tell us about the tail behavior of $\X$? The answer can be conveniently formulated in terms of multivariate generalized Pareto (MGP) distributions, which extend Assumption~\ref{ass:margins_rv} to the multivariate case. These distributions have been introduced in~\cite{rootzen2006} and are reviewed in~\cite{Multivariate-Naveau-segers}. Below, we provide an introduction to their usage in modeling multivariate extremes along with a theoretical result (Proposition~\ref{prop:MGPapprox}) which will be useful for sampling from a MGP distribution.

In arbitrary dimension, it is not clear how to define an ``extreme'' point. Some authors are interested in modeling the vector of componentwise maxima of the data, leading to the multivariate GEV distributions, studied extensively in the literature after the pioneering work of~\cite{dehaan1977}. Recently, see e.g.~\cite{rootzen2018}, more attention has been given to an extension of the standard PoT procedure in a multivariate context. The approach relies on the fact that the excess of $\X$ above a large threshold vector $\u$ asymptotically follows an MGP distribution, with an appropriate definition of when an exceedance over a multivariate threshold takes place.

More concretely, given a vector $\u \in \R^d$ of ``high thresholds'', i.e., below but close to $\u_* = (u_{*j})_{j=1}^d$, where $u_{*j}$ denotes the (possibly infinite) upper endpoint of $X_j$, we consider the random vector of excesses 
\begin{equation}
\label{eq:defYu}
    \Y_{\u} \de \X - \u \mid \X \nleq \u,
\end{equation}
where $\X - \u \de (X_j - u_j)_{j=1}^d$ and where $\nleq$ means that, for at least one $j \in \{1,\ldots,d\}$, we have $X_j > u_j$. Here and below, operations on real numbers are extended to vectors in $\R^d$ in a coordinate-wise manner. Under the aforementioned assumptions, we can establish the asymptotic distribution of $\Y_{\u}$ as $\u \to \u_*$, in a particular way made precise in the statement of the result. 
Weak convergence in $\R^d$ is denoted by $\wc$. 


\begin{proposition}[Weak convergence of excesses to the MGP distribution]
\label{prop:MGPapprox}
Under Assumptions~\ref{ass:multRV}--\ref{ass:margins_rv}, we have
\begin{equation}
\label{eq:XbtatYxi}
    \frac{\X - \bm{b}(t)}{\bm{a}(t)} \, \Big| \, \X \not\leq \bm{b}(t)
     \wc \frac{\bm{Y}^{\bm{\xi}}-1}{\bm{\xi}},
    \qquad t \to \infty,
\end{equation}
where $\bm{Y}$ is a \emph{multivariate Pareto} random vector, with distribution
\begin{equation}
\label{eq:YdYW}
    \Y \stackrel{\mathcal{L}}{=} \lp Y \bTheta \mid Y \bTheta \nleq 1 \rp
\end{equation}
with $Y$ a unit-Pareto random variable independent of $\bTheta \sim \Phi$, and where $\xivec = (\xi_j)_{j=1}^d$ is the vector of marginal coefficients as in Assumption~\ref{ass:margins_rv}; $\bm{a}(\cdot) = (a_j(\cdot))_{j=1}^d$ and $\bm{b}(\cdot) = (b_j(\cdot))_{j=1}^d$ are the same functions as the one introduced in Remark~\ref{rem:equiv_margins}. Consequently, along the curve $\u : t\in (1,\infty) \mapsto \u(t) \de \bm{b}(t)$ we have,
\begin{equation}
\label{eq:wcYu}
    \frac{\Y_{\u(t)}}{\sigvec(\u(t))} \wc \frac{\bm{Y}^{\bm{\xi}}-1}{\bm{\xi}},
    \qquad t \to \infty,
\end{equation}
where $\bm{\sigma}(\cdot) = (\sigma_j(\cdot))_{j=1}^d$ contains the scaling functions in Assumption~\ref{ass:margins_rv}.
\end{proposition}

The proof of Proposition~\ref{prop:MGPapprox} is provided in Appendix~\ref{app:proof_prop1}.

Typically, as for the univariate case, the scaling function in~\eqref{eq:wcYu} is viewed as a scaling parameter and Proposition~\ref{prop:MGPapprox} leads to the approximation in distribution for
\begin{equation}
\label{eq:mgp_approx}
    \Y_{\u(t)} \stackrel{\mathcal{L}}{\approx} \MGP \de \sigvec \frac{\bm{Y}^{\bm{\xi}}-1}{\bm{\xi}}, \qquad t \to \infty
\end{equation}
with $\sigvec \in (0,\infty)^d$ and $\gamvec \in \R^d$ containing, respectively, the scale and shape parameters from the marginal GP distributions in Assumption~\ref{ass:margins_rv} and where $\Y$ is Multivariate Pareto distributed as in~\eqref{eq:YdYW}.

\begin{remark}[Standard MGP random vectors]
The random vector $\MGP$ in \eqref{eq:mgp_approx} is MGP($\sigvec,\xivec,\S$) distributed, in the sense of~\cite[Definition~2.2]{Multivariate-Naveau-segers}, with \emph{spectral generator} $\S$ having distribution
\begin{equation}
\label{eq:spectral_gen}
    \P \lp \S \leq \x \rp = \frac{\E \left[\exp(Q) \, \I\{\U \leq \x + Q\}\right]}{\E[\exp(Q)]}, \qquad \x \in \R^d,
\end{equation}
where $\I\{\mathcal{E}\}$ denotes the indicator random variable of the event $\mathcal{E}$ and $\U \de \log(\bTheta)$, with $\bTheta \sim \Phi$, and $Q \de \max(\U)$. In other words, the random vector $\U$ is a $U$-generator of $\MGP$ in the sense of Proposition~9 of~\cite{rootzen2018multivariate}; see also equation~(12) in~\cite{Multivariate-Naveau-segers}. Note in particular that the moment condition $0 < \E[\exp(U_j)] < \infty$ for all $j \in \{1,\ldots,d\}$ is satisfied as we have, from~\eqref{eq:marginalPhi},
\[
    \E[\exp(U_j)] = \E[\Theta_j] = \int_{\simplex} w_j \, \diff\Phi(\w) = \frac{1}{d}, \qquad j \in \{1,\ldots,d\}.
\]

Here, since we work mainly with the angular measure $\Phi$ in our procedure, we decided to consider the multivariate Pareto random vector $\Y$, as in~\eqref{eq:YdYW}, to be the ``standard'' MGP, that is with $\xivec = \1$ and $\sigvec = \1$ instead of the choice $\xivec = \0$ made in~\cite{rootzen2018multivariate,Multivariate-Naveau-segers, rootzen2018}, for example. This approach has also been considered in~\cite{ferreira2014pareto}. Of course, it suffices to take $\Z = \log(\Y)$ to obtain an MGP random vector with parameter $\xivec = \0$, as can be seen from comparing~\eqref{eq:mgp_approx} to Equation~(4) in~\cite{Multivariate-Naveau-segers}.
\end{remark}

\paragraph{Tail modeling and rare event probabilities.}
Thanks to Proposition~\ref{prop:MGPapprox}, we have now a clear path to modeling the tail of $\X$. Suppose we are interested in estimating a probability $\P(\X \in C)$ for some $C \subseteq \{\x \in \R^d : \x \nleq \u(t) \}$ for some large $t>1$, with the same $\u(t)$ as in~\eqref{eq:wcYu}, so that the approximation of $\Y_{\u(t)} = \X-\u(t) \mid \X \nleq \u(t)$ by the MGP $\MGP$ as in~\eqref{eq:mgp_approx} is accurate. In such a region, there may be no or few observations available, so that an empirical estimate is not reliable. Instead, we write the probability of interest as
\begin{align}
    \P(\X \in C) &= \P(\X \nleq \u(t)) \ \P \lp \Y_{\u(t)} \in C-\u(t) \rp \nonumber \\
    &\approx \P(\X \nleq \u(t)) \ \P \lp \MGP \in C-\u(t) \rp. \label{eq:mgpExtrap}
\end{align}
The first probability can be estimated using an empirical evaluation on the sample while the second one will be computed using an estimate of the MGP distribution of $\MGP$.

As can be seen from~\eqref{eq:mgp_approx}, the MGP random vector $\MGP$ is a simple transformation of the standard MGP random vector $\Y$ introduced in~\eqref{eq:YdYW}, involving only marginal parameters for which estimation is well developed mathematically and numerically. Therefore, the main challenge to obtain informative samples to evaluate tail probabilities related to $\X$ is to obtain realizations of $\Y$. By Proposition~\ref{prop:MGPapprox}, this is possible if we are able to simulate from $\Phi$---at least approximately since it is a limit distribution by nature, so that no direct observation of it will ever be available.

We propose to use a specific GAN structure to achieve this goal. The main principles underlying this framework are recalled in the following section.

\subsection{Wasserstein Generative Adversial Networks}
\label{sec:GANs}

Suppose we are interested in estimating the unknown distribution $\P$ of some, potentially complex, data on a large vector space $\mathcal{X}$. For instance, think of the angular measure $\Phi$ in Section~\ref{sec:GANs} supported on the simplex $\simplex$ for large $d$. Then, the standard parametric approaches based on densities may not be satisfactory as they are often too restrictive to take into account all the possible characteristics of a complex angular measure in large dimension. An alternative direction is to start from a simple random variable $Z \in \mathcal{Z}$ in some ``latent'' space--typically, of lower dimension than $\mathcal{X}$--whose distribution is known and to look for a parametric transformation $g_\theta : \mathcal{Z} \to \mathcal{X}$ such that $g_\theta(Z)$ has a distribution sufficiently close to $\P$. Practically, this has several advantages. We get more flexibility as the map $g_\theta$ can be virtually anything if we consider a sufficiently rich family of parametric functions and the latent space/distribution can also be selected freely. It is easier to simulate from $\P$ as it suffices to obtain a sample of $Z$ (easy) and transform it based on $g_\theta$ rather than more costly algorithms based on the knowledge of the density values.

A Wasserstein GAN (WGAN) architecture~\cite{arjovsky2017} follows this path to estimate $\P$ by considering $g_\theta$ to be a neural network and by measuring the difference in distribution between the real data distribution $\P$ and the law of $g_\theta(Z)$ via the \emph{Earth Mover's Distance (EMD)}, which is a particular instance of the Wasserstein distance in optimal transport~\cite{villani2003}.

A WGAN consists of two neural networks: a so-called ``Generator'' $G = G_\theta : \mathcal{Z} \mapsto \mathcal{X}$ and a ``Discriminator'' $D = D_w : \mathcal{X} \mapsto \R$ (also sometimes called ``Critic'' depending on the authors). These two networks play an opposite role: $G_\theta$ aims to produce samples as realistic as possible and close to the target distribution and $D_w$ aims at efficiently discriminating samples and determining which ones are generated by $G_\theta$ and which ones are real. This ``game'' can be formulated as a min-max optimization problem in the EMD setup that we recall know.

The EMD is simply a distance between probability measures  with finite first moment defined on a metric space $(\mathcal{X}, \rho)$. If $\P$ and $\Q$ are two such distributions, it is equivalently defined as
\begin{align}
    W_1(\P,\Q) 
    &\de \inf_{\pi \in \Pi(\P,\Q)} \int_{\mathcal{X} \times \mathcal{X}} \rho(x,y) \, \diff \pi(x,y) \label{eq:def1EMD} \\
    &= \sup_{f \in \Lip_1} \left\{ \int_{\mathcal{X}} f(x) \, \diff\P(x) - \int_{\mathcal{X}} f(x) \, \diff\Q(x) \right\} , \label{eq:def2EMD}
\end{align}
where $\Pi(\P,\Q)$ is the set of probability measures on $\mathcal{X} \times \mathcal{X}$ with $\P$ and $\Q$ as first and second margins and $\Lip_1$ is the set of functions $f : \mathcal{X} \to \R$ that are $1$-Lipschitz continuous, that is, functions that satisfy
\[
    \sup_{x,y \in \mathcal{X}, x \neq y} \frac{|f(x)-f(y)|}{\rho(x,y)} \leq 1.
\]
If $\mathcal{X}$ is an open subset of $\R^N$ for some $N \in \N$, this implies
\begin{equation}
\label{eq:gradientBound}
    | \nabla f(x) |_2 \leq 1 \qquad \text{for almost every } x \in \mathcal{X},
\end{equation}
that is, for every $x \in \mathcal{X}$ except for a subset of Lebesgue measure zero. The equivalence between~\eqref{eq:def1EMD} and~\eqref{eq:def2EMD} is known as the Kantorovich--Rubinstein duality theorem, the proof of which is to be found in, e.g., \cite[Section~1.2]{villani2003}.

In formulation~\eqref{eq:def2EMD} of the EMD, the discriminator $D_w$ will play the role of the ``separating'' function $f$ and $Q$ will correspond to the measure generated by the generator $G_\theta$, that is, the distribution of $G_\theta(Z)$, while $\P$ will again be our target measure from which we observe a random sample. This leads to the following min-max procedure:
\begin{equation}
\label{eq:min-max}
    \operatorname*{min}_{\theta \in \mathcal{S}_G} \operatorname*{max}_{w \in \mathcal{S}_D} \left\{ \int_{\mathcal{X}} D_w(x) \, \diff\P(x) -  \int_{\mathcal{Z}} D_w \lp G_\theta(z) \rp \, \diff\P_Z(z)  \right\},
\end{equation}
where $\P_Z$ denotes the law of the latent vector $Z$ and $\mathcal{S}_G$ and $\mathcal{S}_D$ denote the (finite-dimensional) parameter spaces of the generator and the discriminator respectively. Optimization is performed in the respective parameter spaces of the generator and the discriminator, which contain all the weights in their respective network structures. Typically, the expectations are replaced by averages over batches of real data with distribution $\P$ and fake data with distribution $\P_Z$. The joint minimization/maximization of the objective function is performed by gradient descent based on the standard backpropagation algorithm. The whole GAN procedure is summarized in Figure~\ref{fig:WGAN}.

\begin{figure}[ht]
    \centering
    \begin{tikzpicture}[
        node distance=1cm and 1cm,
        every node/.style={draw, text width=2cm, align=center, minimum height=1.1cm, font=\small},
        every path/.style={->, thick}
        ]

        \node (noise) {Random Noise $z$};
        \node (generator) [right=of noise] {Generator $G_\theta(z)$};
        \node (critic) [right=of generator] {Critic $D_w(x)$};
        \node (wasserstein) [below=of critic] {EMD $W_1$};

        \node (real) [above=of critic] {Real Data $x$};

        \path (noise) edge (generator);
        \path (generator) edge (critic);
        \path (real) edge (critic);
        \path (critic) edge (wasserstein);

    \draw[<-, red, thick, dashed] (generator.south) 
        to[out=-90, in=180] ++(1.5,-1.2) node[below left, font=\small] {Gradient descent} 
        to[out=0, in=-90] (wasserstein.south);
    
    \draw[<-, red, thick, dashed] (critic.east) 
        to[out=90, in=180] ++(1,1) node[below right, font=\small] {Gradient descent} 
        to[out=0, in=90] (wasserstein.east);

    \end{tikzpicture}
    \caption{Illustration of a Wasserstein GAN with gradient-based optimization}
    \label{fig:WGAN}
\end{figure}

In practice there is no need for the neural network $D_w$ to be Lipschitz continuous as required in definition~\eqref{eq:def2EMD} of the EMD. Initially, in~\cite{arjovsky2017}, the authors proposed to clip the weights, that is, forcing them to lie in a fixed compact space, typically $[-c,c]^{|\mathcal{S}_D|}$ for some small $c>0$. One may show that if the discriminator consists of a multilayer perceptron, which will always be the case here, its Lipschitz norm will be bounded by a finite constant depending only on $c$. Even though this constant may not be equal to one, replacing $\Lip_1$ by $\Lip_C$ for $C > 0$ in~\eqref{eq:def2EMD} only amounts to multiplying the value of the resulting distance by $C > 0$ and it does not play a role in the optimization of the generator.

Since then, it has been shown that in some cases it is preferable to enforce the Lipschitz constraint in a soft manner, using~\eqref{eq:gradientBound}. Of course, controlling the gradient $\nabla D_w$ everywhere on $\mathcal{X}$ is not possible, so what is usually done is to add a penalty term to the objective function of the form
\begin{equation}
\label{eq:gradient_penalty}
    \lambda  \int_{\mathcal{X}} \lp |\nabla D_w(x) |_2 - 1 \rp^2 \, \diff\mu(x), \qquad \lambda > 0,
\end{equation}
where $\mu$ is a probability measure on $\mathcal{X}$. The standard approach in~\cite{gulrajani2017} is to take $\mu$ as the law of the mixture $U \cdot X_r + (1-U) \cdot G_\theta(Z)$ where $X_r \sim \P$ follows the data distribution, $Z \sim \P_Z$ is random noise and $U$ is uniformly distributed on $[0,1]$ and independent of the rest. This choice is motivated by Proposition~1 in~\cite{gulrajani2017} and we follow this approach here.

As explained at the end of Section~\ref{sec:regular-variatio} and in the beginning of this section, our aim is to use this architecture to obtain accurate samples from the angular measure $\Phi$ on $\simplex$. Since the simplex structure of such observations makes their distribution a bit non-standard and could prevent the generator from producing high-quality output, we propose to apply the procedure on a transformation of the angular data to coordinates in an orthonormal basis of the simplex and to transform the coordinates back to angles afterwards. The next section and Appendix~\ref{app:aitchison} provide more details about this transformation.

\subsection{Aitchison coordinates}
\label{sec:aitchison}

A core idea of compositional data analysis~\cite{aitchison1982,Egozcue2003}, mainly developed by John Aitchison (1926--2016), consists of transforming raw simplex data in $\R^d$ to coordinates with respect to an orthonormal basis of a certain $(d-1)$-dimensional subspace of $\R^d$. This makes interpretation  more complicated, but having data on a full linear $(d-1)$-dimensional sample space instead of on the unit simplex in $\R^d$ can improve statistical and neural network modeling substantially. Since we are not all that concerned with interpretation in the WGAN architecture described in Section~\ref{sec:GANs} but instead with obtaining the highest possible  efficiency in  producing accurate new multivariate extreme samples, we will apply this idea in our context. 

Infinitely many orthogonal bases exist for the open simplex $\osimplex$ equipped with its inner product space structure. Here, we work with the one proposed in Proposition~2 of~\cite{Egozcue2003}.
Details of the underlying construction and more information on the Aitchison simplex, i.e., the unit simplex equipped with a certain vector space structure and an inner product, can be found in Appendix~\ref{app:aitchison}. The essence of the matter is to map $\osimplex$ to $\H = \{\x \in \R^d : \sum_{j=1}^d x_j = 0\}$ by means of the centered logratio function ($\clr$) in \eqref{eq:clr}. The set $\H$ is a $(d-1)$-dimensional linear space equipped with the standard Euclidean inner product, the structure of which carries over to $\osimplex$ through the function $\clr$.
\rev{
Recall that for a vector $\x \in \R^d$, $\operatorname{softmax}(\x)$ is the element of $\osimplex$ defined by
\[
	\operatorname{softmax}(\x) \de \lp \frac{\exp(x_j)}{\sum_{i=1}^d \exp(x_i)} \rp_{j=1}^d.
\]
}
\begin{proposition}[An orthonormal basis for the Aitchison simplex]
\label{prop:ONbaseSimplex}
The vectors $\e_1^*,\ldots,\e_{d-1}^*$ defined by
\[
    \e_i^* \de \operatorname{softmax}(e_i) \in \osimplex
    \text{ with }
    e_i \de \sqrt{\frac{i}{i+1}} \lp \underbrace{i^{-1}, \ldots, i^{-1}}_{i \text{ times}}, -1,0 \ldots,0 \rp \in \R^d
\]
for $i \in \{1,\ldots,d-1\}$ form an orthonormal basis of the Aitchison simplex.
\end{proposition}

Now that we have constructed an orthonormal basis of the Aitchison simplex in Proposition~\ref{prop:ONbaseSimplex}, we are able to decompose any vector $\u \in \osimplex$ in this basis as
\[
    \u = \bigoplus_{i=1}^{d-1} \ \langle \u, \e_i^*\rangle_A \odot \e_i^* 
    =  \bigoplus_{i=1}^{d-1} \ \langle \clr(\u), \clr(\e_i^*)\rangle \odot \e_i^* 
    =  \bigoplus_{i=1}^{d-1} \ \langle \clr(\u), \e_i \rangle \odot \e_i^*. 
\]
the notation being defined in Appendix~\ref{app:aitchison}.
The coordinates $\u_i^* \de \langle \clr(\u), \e_i \rangle$ for $i \in \{1,\ldots,d-1\}$ are sufficient to completely describe the vector $\u$. These coordinates will be the quantities entering the WGAN structure described in Section~\ref{sec:GANs} as they typically have a nicer distribution than the original simplex random vectors of interest and hence will be easier to ``learn'', which helps facing the difficulty that sample sizes in extreme value analysis are often not very large.

\section{Combining GAN, Aitchison coordinates, and Extremes}
\label{sec:methodo}

Recall that our main objective is to provide an algorithm which is able to accurately sample from the tail of a random vector $\X \in \R^d$. We do this by combining Algorithm~\ref{alg:WGAN} and Algorithm~\ref{alg:MGP} described below. 
\begin{itemize} 
\item Algorithm~\ref{alg:WGAN} develops a Wasserstein GAN to simulate from the dependence structure of the unit-Pareto transformed observations. 
\item Algorithm~\ref{alg:MGP} uses estimates of the marginal GP parameters to transform these simulated values to the original scale. 
\end{itemize}
The simulated values can then be used to obtain accurate estimates of probabilities of the form $\P(\X \in C)$ where $C \subseteq \{\x \in \EE : \x \nleq \u(t) \}$ for some ``high'' threshold vector $\u(t) \in \R^d$ as in~\eqref{eq:wcYu}, close to $\u_*$.

The main challenge lies in simulation from the angular measure $\Phi$ of  $\V$ in Assumption~\ref{ass:multRV}, since moving from $\Phi$ to the MGP distribution is a matter of scaling and marginal tail estimation, for which theory and practice are well developed. Consequently, we propose to use the WGAN architecture of Section~\ref{sec:GANs} to obtain accurate samples from $\Phi$. This is done via the Aitchison coordinates,  Section~\ref{sec:aitchison}, as detailed in Algorithm~\ref{alg:WGAN} below.

The input to Algorithm~\ref{alg:WGAN} is a set of training observations $\X_1,\ldots,\X_n$, with $\X_i = (X_{i1},\ldots,X_{id})$, which will be 
converted to \chg{approximately unit-Pareto margins} 
using the empirical marginal cumulative distribution functions
\begin{equation}
\label{eq:empiricalCDF}
    \widehat{F}_j(x) \de \frac{1}{n+1} \sum_{i=1}^n \I\{X_{ij} \leq x\},
    \qquad x \in \R, j \in \{1,\ldots,d\}.
\end{equation}
Division by $n+1$ instead of $n$ is there to prevent division by zero in the standardization. The angles associated to ``large'' observations are supposed to provide an approximate sample from $\Phi$ in view of Assumption~\ref{ass:multRV}. The output of Algorithm~\ref{alg:WGAN} is a trained generator $G_\theta$ which is able to produce new coordinates in the Aitchison orthonormal basis, defined by the orthonormal basis $\{\e_1,\ldots, \e_{d-1}\}$ of $\H$, that are hopefully close in distribution to the coordinates of angles associated to the large observations in the training set.

The WA-GAN procedure in Algorithm~\ref{alg:WGAN} only uses the observations for which $R=|\V|_1$ is larger than $t=n/k_1$ so that, by~\eqref{eq:Rtail}, the number of observations used by the algorithm is asymptotically binomial with parameters $n$ and $(dk_1)/n$. Here $k_1$ is a value to be chosen by the user, with larger $k_1$ meaning that more observations are used and hence the variance is smaller, and with smaller $k_1$ (hopefully) making the approximation in Assumption~\ref{ass:multRV} more accurate. Finite sample bounds on the quality of this approximation are given in~\cite{clemencon2023concentration}. The relatively low effective sample size makes it important that the architectures of the generator and the discriminator are simple enough so that their parameters can be learned from a moderate number of examples. Those considerations are further explored in the numerical sections.

Algorithm~2 uses for each margin the $k_2$ largest observations, so that between $k_2$ and $d k_2$ of the multivariate observations are used. According to Remark~\ref{rem:relation_thresholds}, the value of $k_2$ should not be larger than $k_1$. A natural convenient choice is to take $k_1=k_2$.

\begin{algorithm}
\caption{WA-GAN for tail dependence estimation} 
\label{alg:WGAN}
\begin{algorithmic}
\Require observations $\X_1,\ldots,\X_n$, orthonormal basis $\{\e_1, \ldots, \e_{d-1}\}$ of $\H$, and $k_1 \in \{1,\ldots,n\}$ 
\Require gradient penalty coefficient $\lambda > 0$, marginal penalty coefficient $\rho > 0$, the number of discriminator iterations per generator iteration $n_D$, the batch size $m$, Adam hyperparameters $\alpha > 0$ (learning rate), $\beta_1,\beta_2 \in [0,1)$, the latent space dimension $\ell \in \N$, the number of epochs $n_E$
\Require discriminator initial parameters $w_0 \in \mathcal{S}_D$, generator initial parameters $\theta_0 \in \mathcal{S}_G$
\State $t \gets n/k_1$
\For{$i = 1,\ldots, n$}
\State $\hV_i \gets \lp 1 / (1 - \widehat{F}_j(X_{ij}) \rp_{j=1}^d$
\State $R_i \gets |\hV_i|_1$ and $\W_i \gets \hV_i / R_i$ 
\If{$R_i \geq t$}
\State $\W_i^* \gets ( \langle \clr(\W_i), \e_j \rangle )_{j=1}^{d-1}$ 
\EndIf
\EndFor
\State $K \gets \sum_{i=1}^n \I\{R_i \geq t\}$
\State $M \gets (\e_1 \, | \, \ldots \, | \, \e_{d-1}) \in \R^{d \times (d-1)}$
\State $\theta \gets \theta_0$ and $w \gets w_0$
\For{epoch $e = 1, \ldots, n_E$}
\For{$t = 1,\ldots, n_D$}
\For{$i = 1,\ldots, m$}
\State sample $\w_i^*$ from $\{\W_1^*, \ldots, \W_K^*\}$, sample $\z_i$ from a standard multivariate normal distribution on $\R^\ell$, sample $u_i$ from an uniform distribution on $[0,1]$
\State $\widetilde{\w}_i^* \gets G_\theta(\z_i)$
\State $\widehat{\w}_i^* \gets u_i \ \w_i^* + (1-u_i) \ \widetilde{\w}_i^*$
\EndFor
\State $L_D \gets \frac{1}{m} \sum_{i=1}^m \left\{ D_w(\widetilde{\w}_i^*) - D_w(\w_i^*) + \lambda  \lp |\nabla_{\w^*} D_w(\widehat{\w}_i^*) |_2 - 1 \rp^2 \right\}$
\State $w \gets \operatorname{Adam}(L_D, w, \alpha, \beta_1, \beta_2)$ 
\EndFor
\For{$i = 1,\ldots, m$}
\State sample $\z_i$ from a standard multivariate normal distribution on $\R^\ell$
\EndFor
\State $L_G \gets - \frac{1}{m} \sum_{i=1}^m D_w \lp G_\theta(\z_i) \rp + \rho \left|  \frac{1}{m} \sum_{i=1}^m \operatorname{softmax} \lp M  G_\theta(\z_i) \rp - \frac{\1_d}{d} \right|_2^2$
\State $\theta \gets \operatorname{Adam}(L_G, \theta, \alpha, \beta_1, \beta_2)$ 
\EndFor
\State \textbf{Output:} Trained generator $G_\theta$
\end{algorithmic}
\end{algorithm}

Algorithm~\ref{alg:WGAN} is essentially the standard WGAN algorithm with gradient penalty of~\cite{gulrajani2017} applied to the coordinates of the large angles in an orthonormal basis of the Aitchison simplex. The ADAM optimizer which is used to update the weights of the generator and the discriminator at each step is a popular gradient descent algorithm with adaptive learning rate introduced in~\cite{kingma2015}. The function \texttt{Adam()} in the algorithm is one such gradient step and is the default optimizer proposed for the WGAN architecture with gradient penalty. Compared to the standard algorithm, one additional regularization term
\begin{equation}
\label{eq:penalization}
    \rho \left|  \frac{1}{m} \sum_{i=1}^m \operatorname{softmax} \lp M  G_\theta(\z_i) \rp - \frac{\1_d}{d} \right|_2^2,
\end{equation}
where $M$ is the matrix defined in Algorithm \ref{alg:WGAN}, is added to enforce the marginal constraint~\eqref{eq:marginalPhi} in the generator in a soft way and thus to enforce sampling from a genuine angular measure. \rev{The expression in \eqref{eq:penalization} penalizes the (empirical) deviations from the marginal conditions in \eqref{eq:marginalPhi} by forcing the squared Euclidean norm of their differences to be close to zero. Note that $M G_\theta(\z_i)$ is the point in $\H$ with coordinates $G_\theta(\z_i)$ with respect to the orthonormal basis $\{\e_1,\ldots,\e_{d-1}\}$; the $\operatorname{softmax}$ function maps this point to the unit simplex. The regularization term then encourages the empirical distribution of the $m$ points on the unit simplex obtained in this way to the moment constraints in \eqref{eq:marginalPhi}.}

Algorithm~\ref{alg:WGAN} focuses solely on the extremal dependence of $\X$ and is independent of any marginal assumptions besides continuity of the marginal distribution functions. Therefore, if one is interested in computing a quantity which only depends on $\Phi$ in Assumption~\ref{ass:multRV}, such as the coefficients $\theta_J$ introduced in the beginning of Section~\ref{sec:numerical}, this can be done independently of Algorithm~\ref{alg:MGP}. This is not the case for methods such as~\cite{Li-et-al} which directly evaluate sample quality on the scale of the MGP distribution. Note also that even though Algorithm~\ref{alg:WGAN} only considers the angular measure $\Phi$ with respect to the $L_1$-norm, a simple post-processing step allows for any norm on $\R^d$, as explained in Appendix~\ref{app:change_norms}.

\begin{remark}[Standardization of the margins to unit-Pareto]
The procedure in Algorithm~\ref{alg:WGAN} is based on data standardized to unit-Pareto margins, in accordance with Assumption~\ref{ass:multRV}. As in~\eqref{eq:empiricalCDF}, we work with an empirical standardization $\widehat{F}_j$ for each margin $j = 1, \ldots, d$, resulting in a procedure using only the marginal ranks of the original data and solely focusing on dependence, not requiring any assumption on the tail of $F_j$ as in Assumption~\ref{ass:margins_rv}. The same approach was followed in~\cite{einmahl2001nonparametric, einmahl2009maximum} and, more recently, in~\cite{clemencon2023concentration}, for example. An alternative would be to use the empirical distribution function $\widehat{F}_j$ only up to a threshold $u_j < u_{*j}$ and fit a univariate GP distribution above this threshold, relying on Assumption~\ref{ass:margins_rv}. Doing so could improve the quality of the transformation to unit-Pareto margins, but at the price of making Algorithm~\ref{alg:WGAN} also require marginal assumptions. Furthermore, this route has already been explored in~\cite{einmahl1997estimating} and the same authors advocate in~\cite{einmahl2001nonparametric} for the pure nonparametric approach. Therefore, we stick to the rank-based procedure.
\end{remark}

Some care is needed when sampling from $\X \mid \X \nleq \u(t)$ for $t>1$ large, where $\u(t)$ is the vector of thresholds such that for each component separately, the excess probability is $1/t$. For example, a common situation is the case where $\X$ represents some positive multivariate risks, that is, $\X$ takes values in $[0,\infty)^d$. If we directly make use of the formula $\u(t) + \MGP$, with estimated quantities, as suggested by formula~\eqref{eq:mgp_approx}, we risk generating ``extreme'' points for which some coordinates are negative. Our  approach avoids this issue.

\begin{algorithm}
\caption{Sampling from the tail of $\X$} 
\label{alg:MGP}
\begin{algorithmic}
\Require observations $\X_1,\ldots,\X_n$, orthonormal basis $\{\e_1, \ldots, \e_{d-1}\}$ of $\H$, trained generator $G_\theta$, the size of the latent space on which $G_\theta$ has been trained $\ell$, and $k_2 \in \{1,\ldots,n\}$
\Require desired number of samples $n^*$
\For{$j = 1,\ldots,d$}
\State $u_j \gets X_{n - k_2:n,j}$
\State Compute the maximum likelihood estimator $(\widehat{\sigma}_j, \widehat{\xi}_j)$ of a univariate GP distribution
\EndFor
\State $M \gets (\e_1 \, | \, \ldots \, | \, \e_{d-1}) \in \R^{d \times (d-1)}$
\State $i \gets 1$ and $\mathcal{G}$ an empty list of size $n^*$
\While{$i \leq n^*$} 
\State sample $\z$ from a standard multivariate normal distribution on $\R^\ell$
\State $\W \gets \clr^{-1} \lp M  G_\theta(\z) \rp$ 
\State sample $Y$ from a unit-Pareto distribution
\State $\Y \gets Y\W$ 
\If{$\max(\Y) > 1$}
\For{$j = 1,\ldots,d$}
\If{$Y_j \leq 1$}
\State $(\mathcal{G}[i])_j \gets X_{(\lceil n - k_2/Y_j \rceil \vee 1):n,j}$ 
\EndIf
\If{$Y_j > 1$}
\State $(\mathcal{G}[i])_j \gets u_j + \widehat{\sigma}_j (Y_j^{\widehat{\xi}_j} - 1)/\widehat{\xi}_j$ 
\EndIf
\EndFor
\State $i \gets i + 1$
\EndIf
\EndWhile
\State \textbf{Output:} List of $n^*$ random variables $\mathcal{G}$ approximately distributed as $\X \,|\, \X \nleq \u$
\end{algorithmic}
\end{algorithm}

Algorithm~\ref{alg:WGAN} permits to sample from the multivariate Pareto distribution $\Y$ associated to $\X$ using~\eqref{eq:YdYW}. Let $\Y^*$ denote such a generated quantity. From~\eqref{eq:tVVt}, we have
\[
    \Y^* \approxlaw \lp \frac{k}{n} \V \mid \V \nleq \frac{n}{k} \rp,
\]
where $\approxlaw$ means an approximation in distribution. Equivalently,
\[
    \V^* \de \frac{n}{k} \Y^* \approxlaw \lp \V \mid \V \nleq \frac{n}{k} \rp.
\]
Now, to pass from $\V$ to $\X$ one needs to apply the transformation $\bm{b}$ described in Proposition~\ref{prop:MGPapprox}, which is based on the marginal quantile functions. Let $\widehat{\bm{b}}$ denote an estimator of this quantity. The final output of our next algorithm will be
\[
    \X^* = \widehat{\bm{b}}(\V^*) 
    = \widehat{\bm{b}} \lp \frac{n}{k} \Y^* \rp
    \approxlaw \lp \X \mid \V \nleq \frac{n}{k} \rp
    = \lp \X \mid \X \nleq \bm{b} \lp \frac{n}{k} \rp \rp,
\]
by the continuity of  margins. The estimator $\widehat{b}_j((n/k)y)$ for $y>0$ depends on whether $y\leq 1$ or $y > 1$. For $y \leq 1$ we simply use the estimator based on the empirical cumulative distribution function $\widehat{F}_j$. For $y > 1$, the estimator is obtained through Assumption~\ref{ass:margins_rv}, more precisely its equivalent formulation (ii) in Remark~\ref{rem:equiv_margins}.
This leads to
\begin{equation}
\label{eq:bjHat}
    \widehat{b}_j((n/k)y) \de
    \begin{dcases}
        X_{(\lceil n - k/y \rceil \vee 1):n,j}, &\qquad 0 < y \leq 1, \\
        \widehat{b}_j(n/k) + \widehat{a}_j(n/k) \cdot \frac{y^{\widehat{\xi}_j}-1}{\widehat{\xi}_j}, &\qquad y > 1, 
    \end{dcases}
\end{equation}
where $X_{1:n,j} < \ldots < X_{n:n,j}$ are the ascending order statistics in the $j$-th sample, and where $\widehat{a}_j(n/k) \de \widehat{\sigma}_j$ and $\widehat{\xi}_j$ are obtained by maximum likelihood for the GP distribution.

\begin{remark}[Relation between thresholds in both algorithms]
\label{rem:relation_thresholds}
When combining Algorithm~\ref{alg:MGP} with the output of Algorithm~\ref{alg:WGAN}, one actually uses two different thresholds: on the one hand, the threshold $t_1 = n/k_1 > 0$ which has been used to fit the angular measure in Algorithm~\ref{alg:WGAN}, and, on the other hand, the threshold vector $\u(t_2) = \bm{b}(n/k_2)$ for which we aim to sample from $\X \mid \X \nleq \u(t_2)$. As Algorithm~\ref{alg:MGP} relies on the output of Algorithm~\ref{alg:WGAN}, one should ensure that $\u(t_2)$ is large enough so that the approximation of the angular measure $\Phi$ in Algorithm~\ref{alg:WGAN} is valid.
In terms of the random variable $R$ in~\eqref{eq:polar}, it means that we must ensure $R = |\V|_1 > n/k_1$ given that $\X \nleq \u(t_2)$, which is equivalent to $\V \nleq n/k_2$, i.e., $|\V|_\infty > n/k_2$. Since we have $|\V|_1 \geq |\V|_\infty$, it suffices to take $k_2 \leq k_1$. This is illustrated in Figure~\ref{fig:relation_thresholds} for $d=2$.

\begin{figure}[h]
\centering
    \begin{tikzpicture}[scale=3]
    
    \draw[->] (0,0) -- (1.5,0) node[below] {\(V_1\)};
    \draw[->] (0,0) -- (0,1.5) node[left] {\(V_2\)};
    
    \draw[thick] (1,0) -- (0,1);
    \draw[thick] (1,0) -- (1,1) -- (0,1);
    
    \fill[pattern=north east lines, pattern color=gray] (1,1) -- (1,1.5) -- (0,1.5) -- (0,1) -- cycle;
    \fill[pattern=north east lines, pattern color=gray] (1,1) -- (1,0) -- (1.5,0) -- (1.5,1.5) -- (1,1.5) -- cycle;

    \node at (0.3,0.3) {\(|\V|_1 = \frac{n}{k_1}\)};
    \node at (0.8,1.1) {\(|\V|_\infty = \frac{n}{k_1}\)};
    
    \end{tikzpicture}
\caption{Relation between thresholds in the $\V$ space. The gray zone is where we can safely sample using Algorithm~\ref{alg:MGP}.}
\label{fig:relation_thresholds}
\end{figure}
\end{remark}

\section{Numerical experiments}
\label{sec:numerical}

We use both simulated and real data sets to explore the performance of our proposed method and to compare it to other existing methods. We start with providing the details of how the different methods are compared and how the  the computations are made. The next section uses simulated data to compare WA-GAN with HTGAN and GPGAN. In the final section these methods are applied to a financial data set.

\subsection{Performance and computation}
\label{sec:numericalmethods}

\paragraph{Performance measures.}
To assess the quality of the generated sample $\X^\gen$, we will typically compare it with a test set $\X^\test$, which is a part of the data that has not been used for training. We propose to use measures that are inspired by multivariate extreme value theory since our focus is on the tail of $\X$.

 A first set of measures that we propose is based on the \emph{extremal coefficients} ~\cite[Section~8.2.7]{beirlant2004}. For any non-empty subset $J \subseteq \{1,\ldots, d\}$ with at least two elements $|J| \geq 2$, we define its extremal coefficient $\theta_J$ as
\begin{equation}
\label{ex:extCoefJ}
    \theta_J \de d \cdot \int_{\simplex} \bigvee_{j \in J} w_j \ \diff\Phi(\w) \, \in [1, |J|].
\end{equation}
The closer $\theta_J $ is to $1$, the stronger the dependence in the tail of $(X_j)_{j \in J}$, while $\theta_J$ close to $|J|$ corresponds to weaker tail dependence. For a given coefficient order $k = |J|$, to compute a summary score of their difference in the test set $\X^{\test}$ and the generated set $\X^\gen$, we compute their relative absolute difference as follows:
\begin{equation}
\label{eq:extCoefScore}
    E_{\overline{\theta}}(k) \de \frac{1}{\binom{d}{k}} \sum_{\substack{J \subseteq \{1,\ldots,d\} \\ |J| = k}} \left| 1 - \frac{\theta_J^\gen}{\theta_J^\test} \right|.
\end{equation}
The coefficients are estimated for the test set using the empirical angular measure in \eqref{ex:extCoefJ} and for the generative methods using the sampled realizations of $\Phi$. For the test set, the radial threshold $n/k$ is the same as for the training set.
Typically, we will consider the bivariate coefficients, with $k = 2$, as in~\cite{Boulagiem-et-al}, and the trivariate coefficients, with $k = 3$. This measure focuses on the extremal dependence structure only and aims to measure the quality of Algorithm~\ref{alg:WGAN}.

A second measure that we propose, which also focuses on the tail of $\X$, aims at evaluating the output of Algorithm~\ref{alg:MGP}. Let $\X^\test$ denote a set of observations distributed like $\X$, which typically have not been used for training, and let $\X^\gen_{\u}$ denote a set of observations generated by Algorithm~\ref{alg:MGP} based on the threshold vector $\u = \bm{b}(n/k_n) \in \R^d$. Then if the MGP approximation is accurate enough, we expect $\X_{\u}^\test \de \X^\test \,|\, \X^\test \nleq \u$ to be close in distribution to $\X^\gen_{\u}$. Hence, we propose to compute their difference in distribution using a simple extension of the EMD in Section~\ref{sec:GANs} which is known as the $2$-Wasserstein distance~\cite[Chapter~7]{villani2003}. If $n_\gen$ and $n_\test$ denote, respectively, the size of the test set and the generated set of observations, the measure can be expressed as
\begin{equation}
\label{eq:FIDscore}
    W_2(\X_{\u}^\gen, \X^\test_{\u}) \de 
    \inf_{\pi \in \Pi \lp \tfrac{\1_{n_\gen}}{n_\gen}, \tfrac{\1_{n_\test}}{n_\test} \rp} \sqrt{\sum_{i = 1}^{n_\gen} \sum_{j = 1}^{n_\test} \|(\X_{\u}^\gen)_i - (\X_{\u}^\test)_j \|_2^2 \, \pi_{ij}}
\end{equation}
where, for (probability) vectors $\bm{a} \in \R^k$ and $\bm{b} \in \R^\ell$, we use the notation  
\[
    \Pi(\bm{a},\bm{b}) \de \left\{ \pi \in \R^{k \times \ell}_+ : \pi \1_\ell = \bm{a} \text{ and } \pi^T\1_k = \bm{b} \right\}.
\]
We can view $\pi_{ij}$ as the amount of mass transferred from $(\X_{\u}^\gen)_i$ to $(\X_{\u}^\test)_j$.
In a similar way as in definition~\eqref{eq:def1EMD} of the EMD, $W_2(\X_{\u}^\gen, \X^\test_{\u})$ measures, among all the possible ways to transport the empirical distribution of the vectors in $\X_{\u}^\gen$ to the one of the vectors in $ \X^\test_{\u}$, the cheapest way to do so if the cost of unitary transportation is equal to the squared Euclidean distance between points.
In the case where we would have normally distributed quantities, this 
quantity would reduce to computing the well-known \emph{Fréchet inception distance}, which is popularly used to assess the quality of generated images~\cite{heusel2017fid} using generative systems like the GAN, also used in~\cite{Li-et-al} to assess the quality of the MGP output of their algorithm. As we compare quantities that are far from being normally distributed, we think it makes more sense to compute the $2$-Wasserstein distance instead.

\paragraph{Architectures.}
\label{sec:architectures}

For both the generator and the discriminator in Algorithm~\ref{alg:WGAN}, we consider fully connected multilayer perceptrons with leaky rectified linear unit activation function~\cite{maas2013rectifier} which applies the map
\begin{equation}
\label{eq:leaky}
    x \in \R \mapsto
    \begin{cases}
        x, &\text{if } x \geq 0, \\
        \alpha x, &\text{otherwise},
    \end{cases}
\end{equation}
to each component of the linear transformation in a given layer, where $\alpha \in [0,1]$ is  typically set to $0.01$. Taking $\alpha = 0$ leads to the rectified linear unit activation function and $\alpha = 1$ leads to a simple linear layer.
We refer to~\cite{maas2013rectifier} for a comparison with other standard activation functions. Here, we consider $\alpha = 0.01$ for any hidden layer while we take $\alpha = 1$ for the final layer in each network. More recently, $\alpha$ has been considered as a learnable parameter in the optimization process~\cite{he2015delving} but this option is not considered in this work. Other architectures are possible if the data have some structure. For example, the authors of~\cite{Boulagiem-et-al} consider convolutions in their network as they work with spatial data on a grid, and this could be adapted to our framework too, but here we only consider basic multivariate data and keep this extension for future work.

\paragraph{Hyperparameter search.}

In addition to the learnable parameters of the generator and discriminator networks in Algorithm~\ref{alg:WGAN}, many external parameters (hyperparameters) can be chosen by the user. We consider the following values for the hyperparameters:
\begin{itemize}
    \item batch size $m \in \{\lfloor n/k \rfloor : k = 1,2,4,8,16\}$;
    \item dimension of the generator's latent space $\ell \in \{\lfloor (d-1) \cdot k \rfloor : k = 0.25, 0.5, 0.75, 1, 2\}$;
    \item maximum number of neurons in a hidden layer in $\{32, 64, 128, 256, 512\}$;
    \item number of hidden layers  in $\{1,2,4,8\}$;
    \item learning rate for the ADAM optimizer in $\{0.01, 0.005, 0.001, 0.0005, 0.0001\}$ and coefficients $\beta = (\beta_1,\beta_2)$ used for computing running averages of gradient and its square in $\{(0.0, 0.9), (0.5,0.9), (0.5,0.99)\}$;
    \item the gradient penalty coefficient $\lambda \in \{0.1,1,3,5,7,9\}$;
    \item the marginal penalty coefficient $\rho \in \{0.001,0.01,0.1,1,3\}$;
    \item the number of discriminator iterations per generator iterations $n_D \in \{1,3,5,10\}$.
\end{itemize}
Inside the space of possible values for those hyperparameters, we perform a random search. Typically, we consider around $2000$ different models in this space for a given dataset. We train them with $n_e = 5000$ epochs, since training them even longer did not significantly improve the performance, but greatly affects the computation time. Evaluation of the associated models is based on the extremal coefficient metric $E_{\overline{\theta}}(k)$ in~\eqref{eq:extCoefScore} with $k = 2$ and $k=3$. \rev{The validated model is chosen to be the one with the lowest value of $\lp E_{\overline{\theta}}(2) + E_{\overline{\theta}}(3) \rp /2$ among the $2000$ models. This validation is performed on a separate dataset from the one used for training or testing.}

\paragraph{Comparison with existing methods.} 

As reviewed in Section~\ref{sec:intro}, numerous methods already exist in the rapidly growing field that combines generative approaches with extreme value theory, each with its own advantages and limitations. Therefore, an exhaustive comparison with all existing methods is not feasible, even less so as the software underlying the numerical results is often not available from the papers. Instead, we choose to compare our approach with versions of two recent methods: HTGAN from~\cite{Girard-et-al} and GPGAN from~\cite{Li-et-al}. We discuss our specific implementations in Appendix~D in the supplement.

\paragraph{Software.} The code used to produce the results of this section was written in \texttt{Python}---in particular, the \texttt{PyTorch}~\cite{PyTorch2019} framework was used to train the generative models---and in \texttt{R}~\cite{R2023}, which was mainly used for data preprocessing and computation of the performance measures associated with the generated data, notably via the \texttt{transport}~\cite{Rtransport2024} package for the computation of the Wasserstein distance $W_2$ in~\eqref{eq:FIDscore}.

\paragraph{Hardware.} The training of the various algorithms and the computation of performance metrics have been performed on the CPUs of the Lemaitre4 and the NIC5 computer clusters, which are managed by the Consortium of Équipements de Calcul Intensif (CÉCI). Details and more specifications related to these computers can be found on the CECI website \url{https://www.ceci-hpc.be/}.

\subsection{Simulated data}
\label{sec:simus}

\bgroup
\color{changecol}

As an illustration of the method’s performance in a controlled setting, we begin by considering simulated data. In Section~\ref{sec:simus:logistic}, we adopt the same experimental setup as in Section~3.2 of~\cite{Girard-et-al}, with which we compare our method. The dependence structure is logistic, characterized by a single parameter that controls the strength of dependence. This scenario is particularly simple in the sense that each sub-vector exhibits the same form of dependence, and increasing the dimension does not necessarily make the problem more difficult, since the algorithm can easily learn from this highly symmetric structure.
For this reason, we consider in Section~\ref{sec:simus:HR} a Hüsler--Reiss dependence structure, where each of the $d(d-1)/2$ pairs of random variables in $\X$ may have a different level of dependence. Finally, in Section~\ref{sec:simus:gaussian}, we turn to an asymptotically independent copula—the Gaussian copula—as the dependence structure of $\X$. In this case, Assumption~\ref{ass:phi_interior} is no longer satisfied, since the angular measure $\Phi$ is concentrated on the vertices of the simplex. Here, our interest lies in assessing whether our methodology can be improved by allowing for a heavier-tailed latent distribution in the generator.

\egroup

\subsubsection{Logistic dependence structure}
\label{sec:simus:logistic}

In this setup, the random vector $\X$ is generated  using a $d$-dimensional Gumbel copula with parameter $\theta \in [1,\infty)$ and Pareto distributed marginals with parameter $\alpha > 0$. It is easy to verify that this data-generating process satisfies Assumptions~\ref{ass:multRV} to \ref{ass:margins_rv} in Section~\ref{sec:regular-variatio}. In particular, the extremal coefficient introduced as a performance measure at the beginning of Section~\ref{sec:numerical} is $\theta_J = |J|^{1/\theta}$. Hence, $\theta \to 1$ corresponds to less dependence in the tail $\X$, while $\theta \to \infty$ corresponds to perfect tail dependence.

We consider various scenarios in which we assess the performance of WA-GAN and compare it to HTGAN and GPGAN. For HTGAN we follow \cite{Girard-et-al} and make the unrealistic assumption that $\alpha$ is known. We take $d \in \{10,20,50\}$ and $\theta \in \{4/3,2,4\}$ leading to a Kendall's tau of $\tau \in \{1/4,1/2,3/4\}$. For the margins, we consider $\alpha = 2$ in any scenario. The models are trained on $n_{\text{train}} = 10\,000$ observations and validated on $n_{\text{val}} = 5\,000$ observations. The performance assessment is done on $n_{\text{test}} = 20\,000$ points.

\paragraph{Results.}

For each method and each scenario, two scores are computed. On the one hand, a dependence score equal to $\{E_{\overline{\theta}}(2) + E_{\overline{\theta}}(3)\} / 2$, with $E_{\overline{\theta}}(k)$ as in~\eqref{eq:extCoefScore}, is used to validate the hyperparameters and to assess the performance of the tail dependence in the generated data. On the other hand, to assess the quality of generated extremes $\X \mid \X \nleq \u$, the $2$-Wasserstein distance as defined in~\eqref{eq:FIDscore} is computed, where $u_j = \widehat{b}_j(n/k_2)$ with $\widehat{b}_j$ as in~\eqref{eq:bjHat}, for some $k_2 \in \{1,\ldots,n\}$ which satisfies $k_2 \leq k_1$ where $k_1 \in \{1,\ldots,n\}$ is used to train the generator in Algorithm~\ref{alg:WGAN}. Here we take $n = n_{\text{train}}$ and $k_1 = k _2 = [\, \sqrt{n} \,]$.

The results are reported in Table~\ref{tab:gan_scores} below.

\begin{table}[h]
\centering
\begin{tabular}{ll|ccc|ccc}
\toprule
 & & \multicolumn{3}{c|}{Dependence score} & \multicolumn{3}{c}{Extremes score} \\
$\tau$ & Model & $d=10$ & $d=20$ & $d=50$ & $d=10$ & $d=20$ & $d=50$ \\
\midrule
\multirow{3}{*}{$\tau = \frac{1}{4}$} 
& WA-GAN &  $\bm{0.0103}$ & $\bm{0.0074}$ & $\bm{0.0054}$  & 67.311 & $\bm{62.905}$ & $\bm{42.825}$  \\
& HTGAN  &  0.0146 & 0.0185 & 0.0199  &  1479.9 & 3106.7 & 1170.7  \\
& GPGAN  & 0.0262 & 0.0321 & 0.0326  &  $\bm{66.663}$ & 65.429 & 52.133  \\
\midrule
\multirow{3}{*}{$\tau = \frac{1}{2}$} 
& WA-GAN &  $\bm{0.0176}$ & 0.0207 & $\bm{0.0097}$  &  $\bm{66.734}$ & $\bm{52.832}$ & $\bm{32.468}$  \\
& HTGAN  &  0.0247  & $\bm{0.0179}$ & 0.0126 &  2725.3 & 1447.9 & 2367.4  \\
& GPGAN  &  0.0511 & 0.0671 & 0.0672  & 67.841 & 55.308 & 46.996  \\
\midrule
\multirow{3}{*}{$\tau = \frac{3}{4}$} 
& WA-GAN &  $\bm{0.0208}$ & $\bm{0.0266}$ & $\bm{0.0158}$  &  $\bm{59.548}$ & $\bm{44.602}$ & $\bm{31.808}$  \\
& HTGAN  &  0.0284 & 0.0338 & 0.055  &  891.62 & 612.44 & 440.45  \\
& GPGAN  &  0.0497 & 0.0726 & 0.0818  &  59.919 & 53.758 & 45.541  \\
\bottomrule
\end{tabular}
\caption{Logistic dependence structure. Dependence and extremes scores for different dependence $\tau$ and dimensions $d$. For each scenario and each score, the best score (i.e., the lowest) is indicated in bold.}
\label{tab:gan_scores}
\end{table}

These results show that WA-GAN is competitive with other methods, from the tail dependence perspective as well as from the quality of the generated extremes. The advantage of WA-GAN seems to increase as the dimension $d$ increases. Large values of $d$ is often at the center of interest when generative models for the dependence are used.
Note also that it may seem surprising at first that the Wasserstein scores improve as $d$ increases. This can be explained by several factors. One possible explanation in the present setting is that the dependence structure of the data-generating process is identical across all pairs of variables, which makes it easier to learn as the dimensionality of the space increases. Consequently, to ensure that the comparison remains meaningful, the main interest lies in comparing the values for a given dimension.

Some additional comments can be made for each method. For WA-GAN, the penalization coefficient $\rho$ in~\eqref{eq:penalization} is typically selected to be large among the possible values ($\rho=1$ or $\rho=3$) showing that it is a good idea to enforce marginal constraints~\eqref{eq:marginalPhi} in the generative model. For HTGAN, the obtained values of the hyperparameters match the discussion on pages 13--15 in the paper~\cite{Girard-et-al} proposing this method. In particular, we get big batch sizes, big latent space dimensions, and low values of the learning rate included in the range recommendend by the authors. For GPGAN, the training phase is quicker since, as illustrated on Figure~\ref{fig:relation_thresholds}, we have a lower training size in this case as this method is trained on points with $|\V|_\infty > n/k$ while WA-GAN is trained on points with $|\V|_1 > n/k$, which is less restrictive. This can be an asset if we have enough data, but can harm GPGAN's performance if $d$ is large and $n$ is not too big.

\begin{figure}[h]
        \centering
        \includegraphics[width=0.49\linewidth]{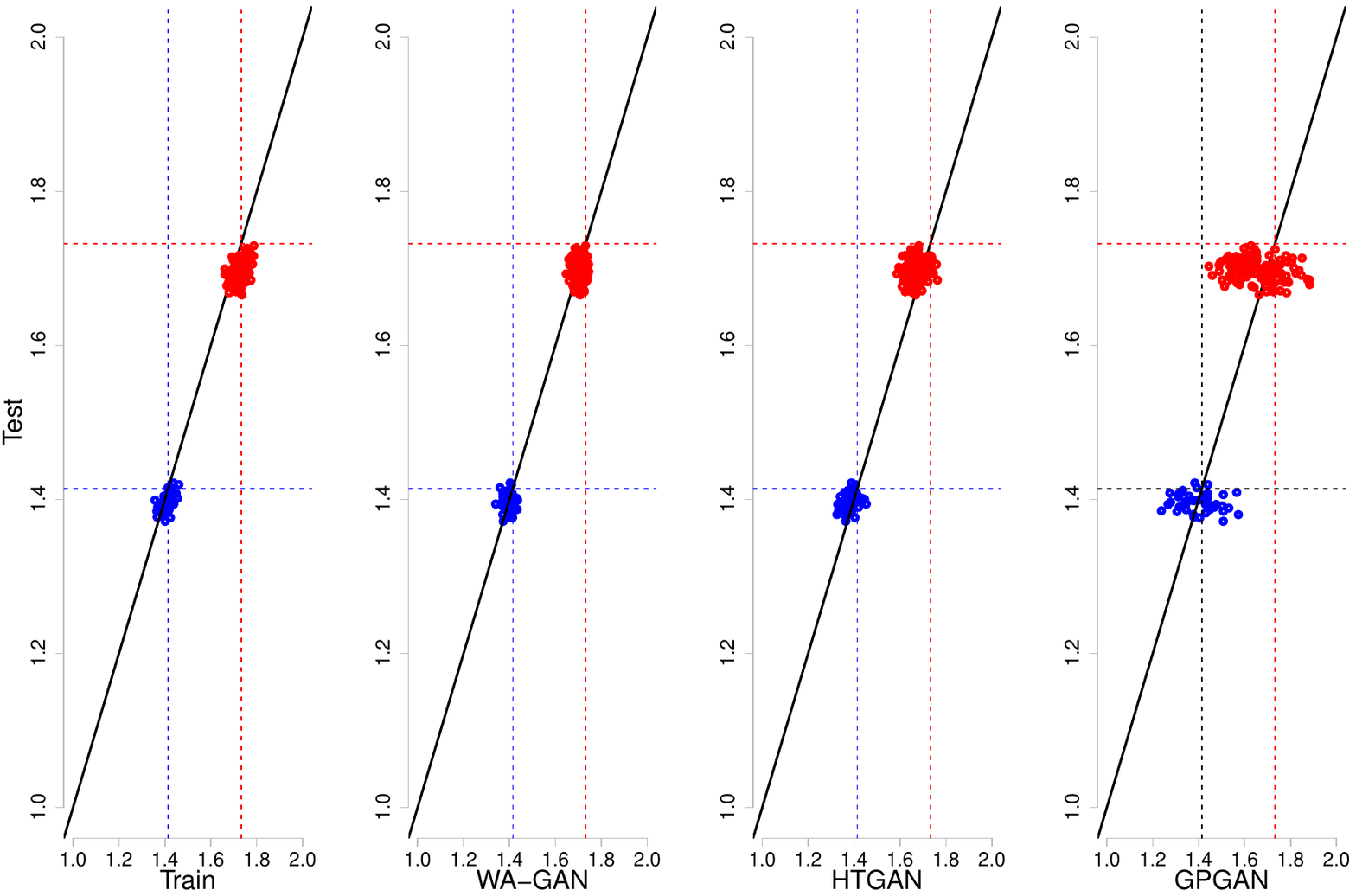}
        \includegraphics[width=0.49\linewidth]{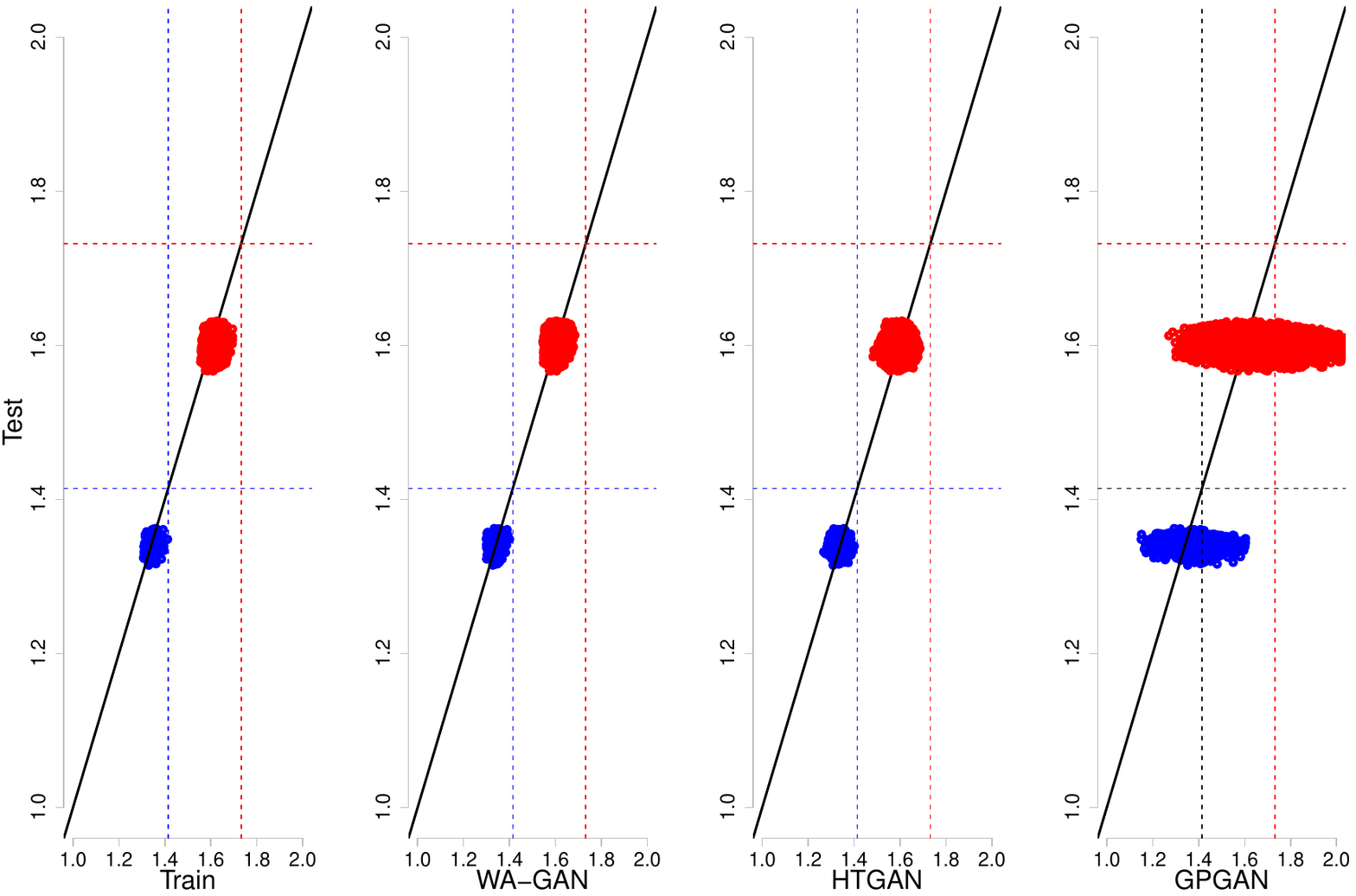}
        \caption{Logistic dependence structure. Left: $d=10$; right: $d=50$. -- Extremal coefficients of order $k=2$ (blue) and $k=3$ (red). Solid black line is the diagonal. Dashed blue and red lines are the true values of the coefficients for $k=2$ and $k=3$ respectively.}
        \label{fig:d=1050}
\end{figure}

As an illustration of the performances of the different methods to accurately reproduce the tail dependence of the data, we plot the empirical bivariate and trivariate extremal coefficients ($|J| = 2$ and $|J| = 3$ in~\eqref{ex:extCoefJ}) of the test set against the ones of the training set, and against those of all the generative methods for $d \in \{10,50\}$ and $\tau = 0.5$. In this setting, we know that the true values of these coefficients are $\sqrt{|J|}$. This is displayed n Figure~\ref{fig:d=1050} for $d \in \{10,50\}$. There is no clear visual difference between WA-GAN and HTGAN, both performing well at representing the coefficients of the test set. GPGAN, however, fails to do so, and even more when the dimension increases. When $d$ gets larger, the empirical extremal coefficients are below their true values, even in the test set. The reason is that the empirical estimator of $\Phi$ in~\eqref{ex:extCoefJ} becomes less accurate as $d$ increases~\cite{clemencon2023concentration}.

\begin{figure}[h!]
        \centering
        \includegraphics[width=0.49\linewidth]{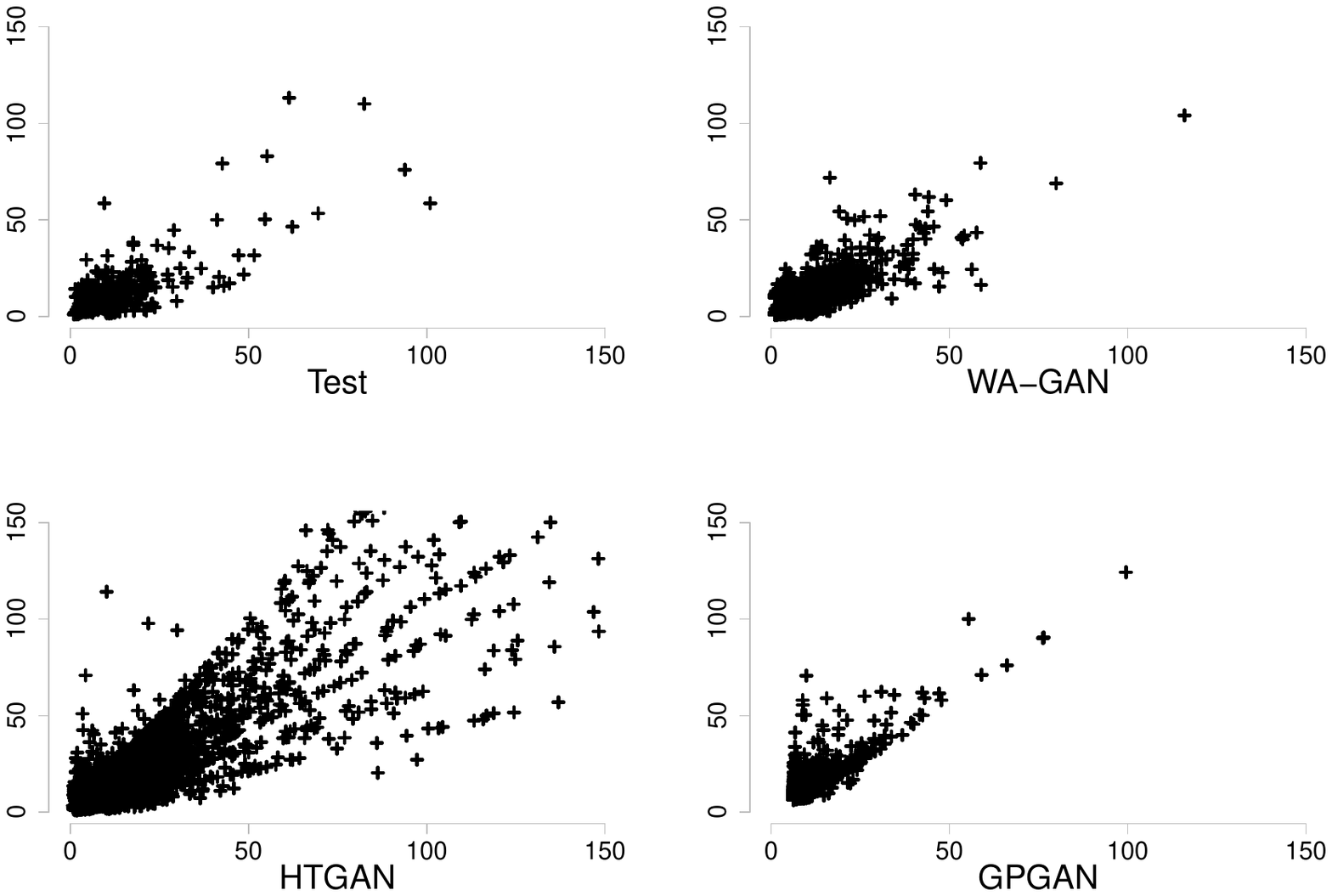}
        \includegraphics[width=0.49\linewidth]{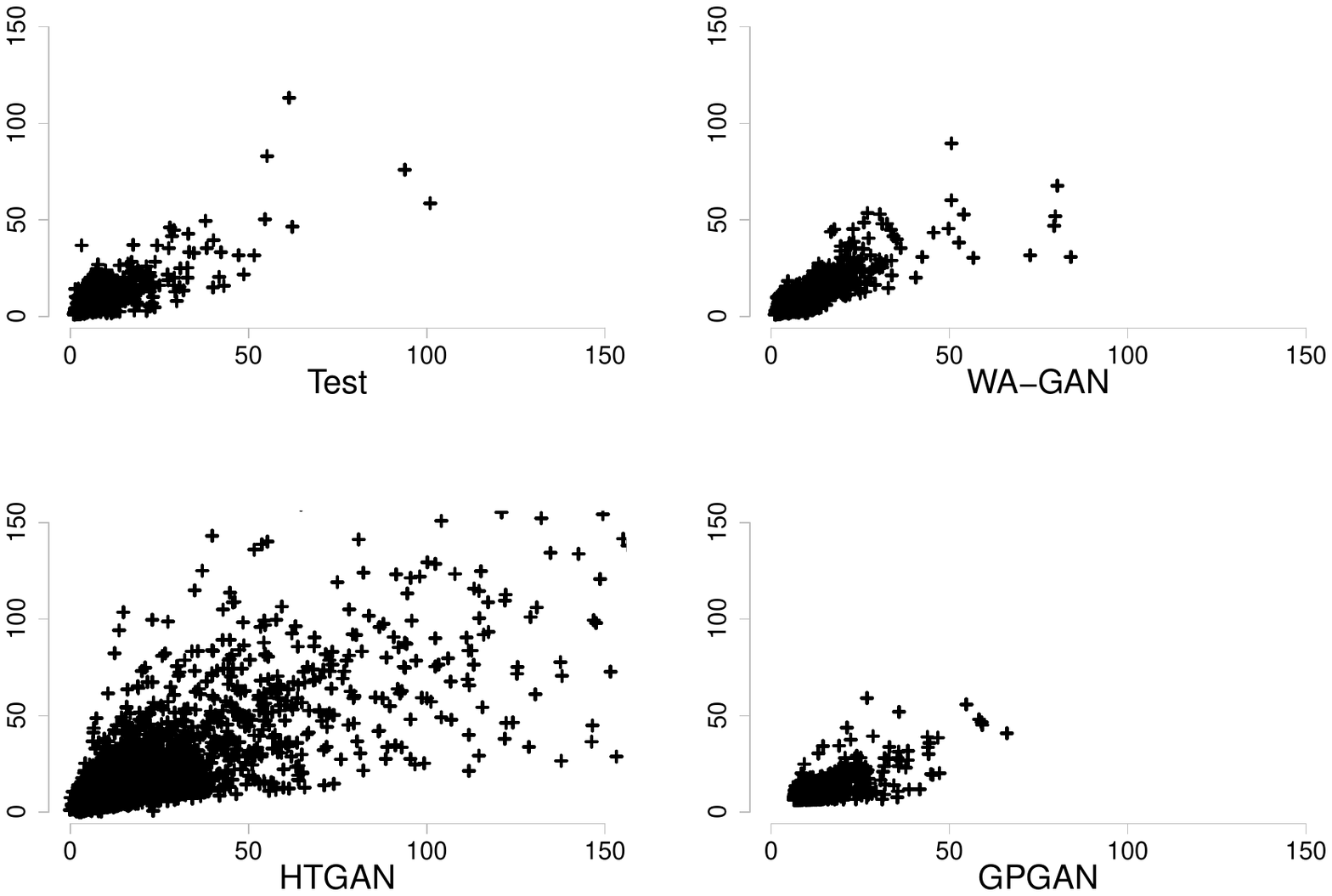}
        \caption{Logistic dependence structure. Left: $d=10$; right: $d=50$. -- First two margins of extremes in the test set and the different generative methods.}
        \label{fig:d=1050E}
\end{figure}

Next, to illustrate the accuracy of generated extremes by each method, we plot the projection of each generated sample on the first two margins and compare it to the first two margins of the extremes in the test set. We take again $\tau = 0.5$. Results for are displayed on Figure~\ref{fig:d=1050E}. WA-GAN seems to perform well, whatever the dimension. The results of HTGAN are less satisfying as it is clear from the figure that the angular measure associated with this method is discrete (this corresponds to the ``directions'' that we observe in the generated extremes) and this does not match the true angular measure which is logistic. This is theoretically justified by Proposition~2.10 in~\cite{Girard-et-al} introducing this method, and already visible from their Figure~2 on page~14. GPGAN also performs well even though its performance decreases in comparison to WA-GAN as the dimension increases; see Table~\ref{tab:gan_scores}.

\bgroup
\color{changecol}

\subsubsection{Hüsler--Reiss dependence structure}
\label{sec:simus:HR}

In this setup, the random vector $\X$ is generated using a $d$-dimensional multivariate Pareto distribution with Hüsler--Reiss dependence structure having a random parameter matrix and Pareto(2) margins.
The parameter matrix generation and the random sampling are done using the \texttt{R} package \texttt{graphicalExtremes}~\cite{graphicalExtremes}. The family is parametrized by a variogram matrix $\Gamma = (\gamma_{ij})_{i,j=1}^d$ where $\gamma_{ij} = 0$ means complete dependence and $\gamma_{ij} = +\infty$ means asymptotic independence.

We consider various scenarios in which we assess the performance of WA-GAN and compare it to GPGAN, whose performance in generating multivariate extremes, according to the previous section, is quite close to the results of our procedure. The models are trained on $n_{\text{train}} = 10\,000$ observations and validated on $n_{\text{val}} = 5\,000$ observations. The performance assessment is done on $n_{\text{test}} = 20\,000$ points.

To illustrate the variety of tail strength dependence in the different scenarios, we plot the parameter matrix of each case as heatmaps in Appendix~E in the supplement.

\paragraph{Results.} The setup and the meaning of the scores are exactly the same as explained above Table~\ref{tab:gan_scores}.

\begin{table}[h]
\centering

\begin{tabular}{l|ccc|ccc}
\toprule
& \multicolumn{3}{c|}{Dependence score} & \multicolumn{3}{c}{Extremes score} \\
Model & $d=10$ & $d=20$ & $d=50$ & $d=10$ & $d=20$ & $d=50$ \\
\midrule
WA-GAN & $\bm{0.0145}$ & $\bm{0.0176}$ & $\bm{0.0123}$ & $\bm{16.69}$ & $\bm{9.512}$ & $\bm{14.926}$  \\
GPGAN  & 0.0629 & 0.1431 & 0.0989  & 20.979 & 11.406 &  21.067 \\
\bottomrule
\end{tabular}

\caption{Hüsler--Reiss dependence structure. Dependence and extremes scores for different dimensions $d$. For each dimension and each score, the best score (i.e., the lowest) is indicated in bold.}
\label{tab:gan_scores_hr}
\end{table}

WA-GAN remains highly competitive in this more complex scenario, both in terms of dependence and generated extremes. Even though the degrees of dependence vary considerably in this setting (Figure~\ref{fig:d=1050_HR}), WA-GAN is able to capture them quite accurately, even in high dimensions. For GPGAN, the situation is less favorable, as the deviation from the reference coefficients becomes quite pronounced when $d$ increases.
\begin{figure}[h]
        \centering
        \includegraphics[width=0.49\linewidth]{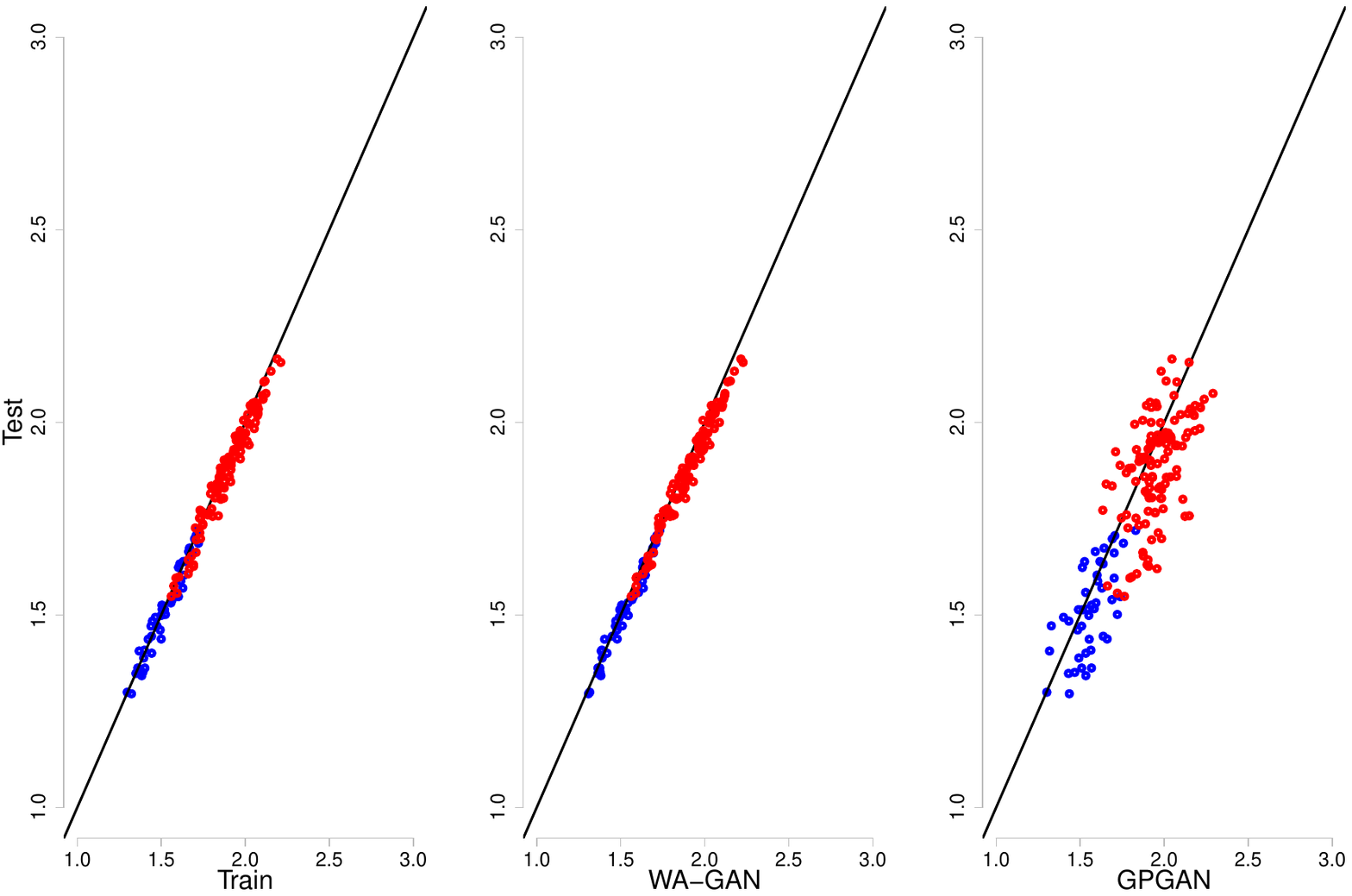}
        \includegraphics[width=0.49\linewidth]{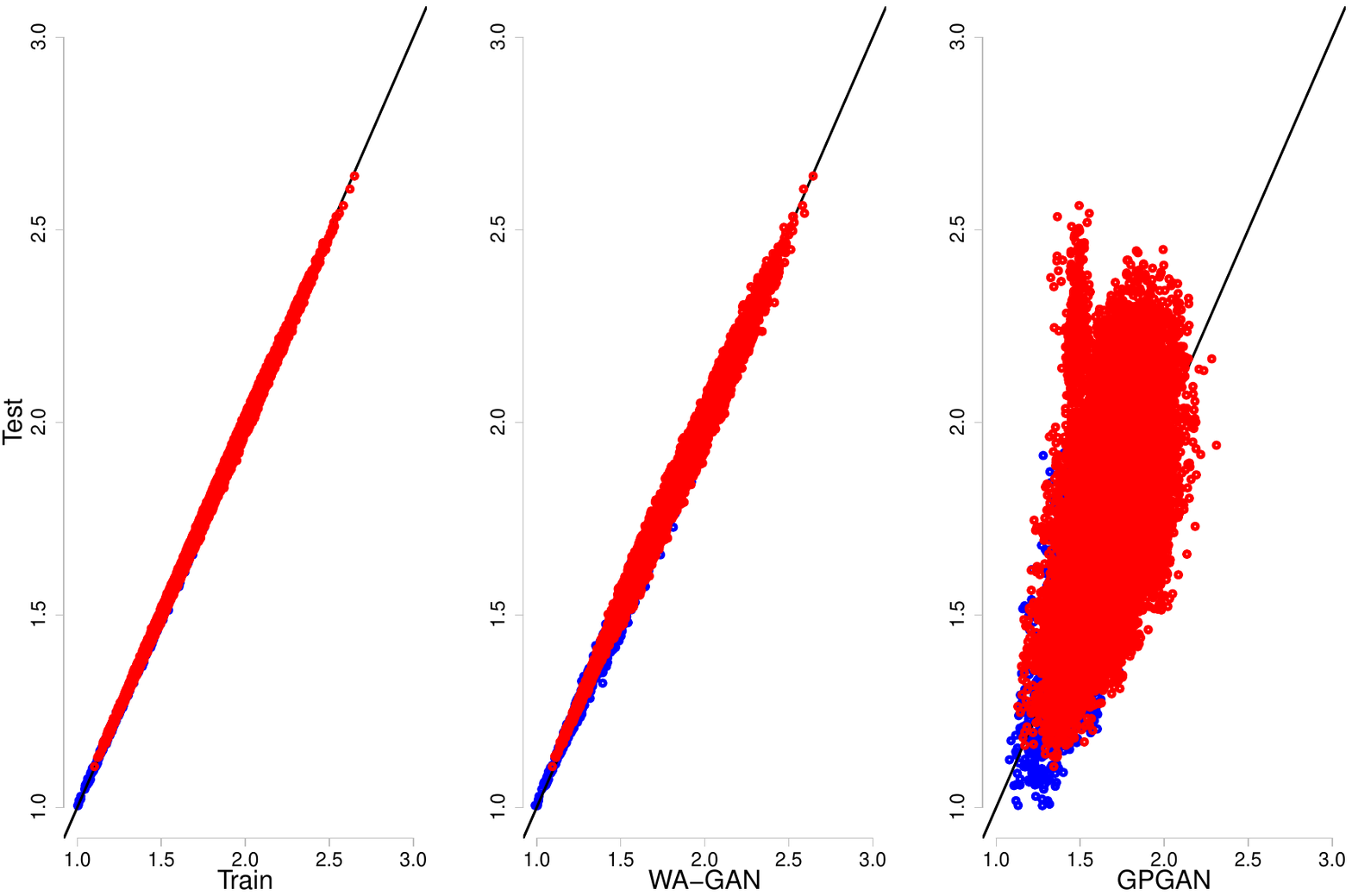}
        \caption{Hüsler--Reiss dependence structure. Left: $d=10$; right: $d=50$. -- Extremal coefficients of order $k=2$ (blue) and $k=3$ (red).}
        \label{fig:d=1050_HR}
\end{figure}

We investigate WA-GAN's sensitivity to the assumption that the angular measure is concentrated on $\osimplex$. Indeed, by construction, the method does not allow the simulation of extremes associated with directions on the boundary of the simplex $\simplex$. Based on the parameter matrices shown in Figure~1 in Appendix~E, we consider, for $d \in \{10,50\}$, two scenarios exhibiting high and low tail dependence between two components, and we compare the generated extremes to the test extremes in Figure~\ref{fig:extremes-HR}. WA-GAN performs well when there is moderate to strong tail dependence but fails to generate extremes that are too close to the axes, corresponding to the low tail dependence scenario, especially if $d=50$. We discuss this issue in the next section under the asymptotic independence scenario and propose a possible improvement to the method.
\begin{figure}[h]
        \centering
        \includegraphics[width=0.49\linewidth]{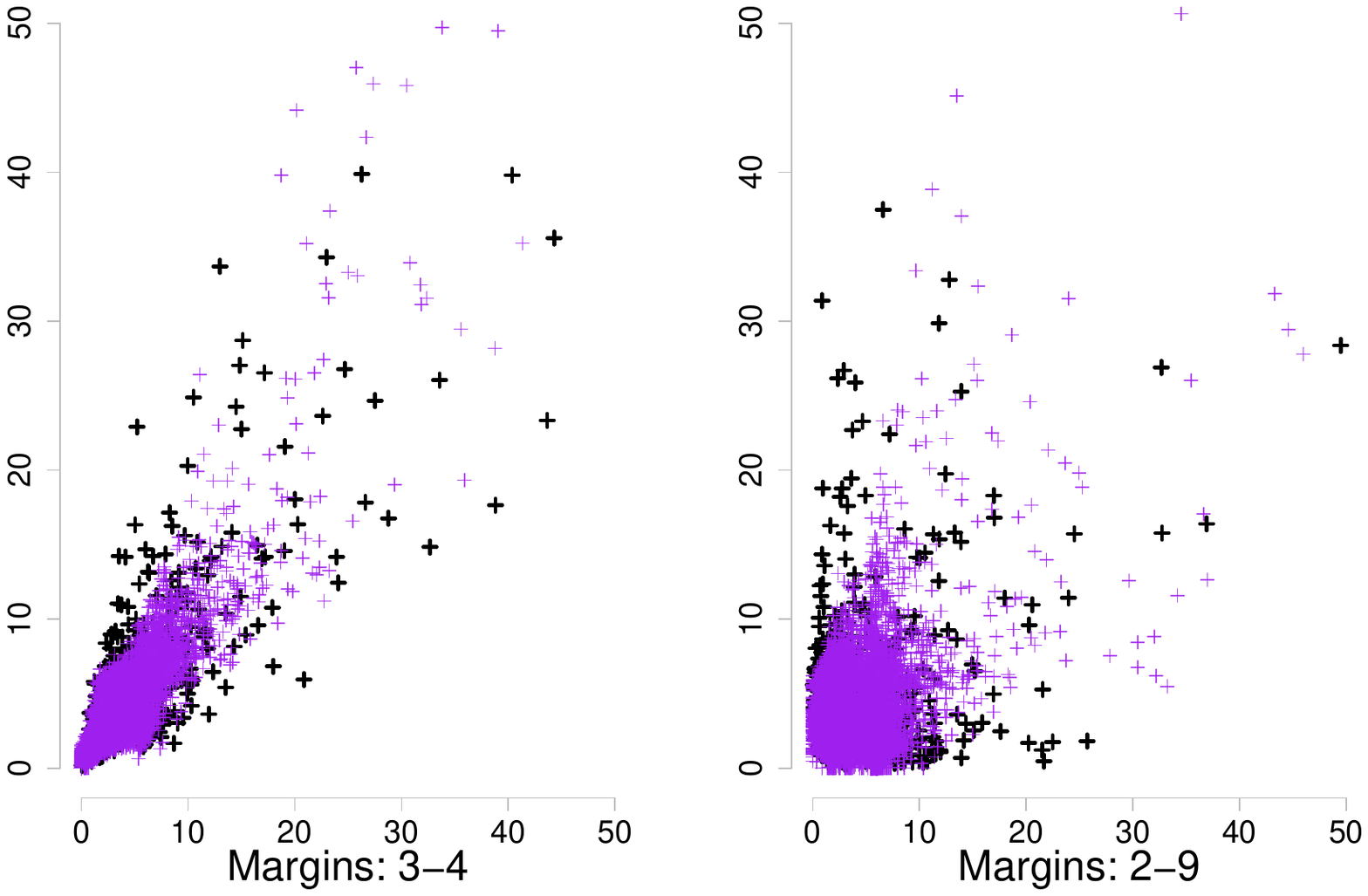}
        \includegraphics[width=0.49\linewidth]{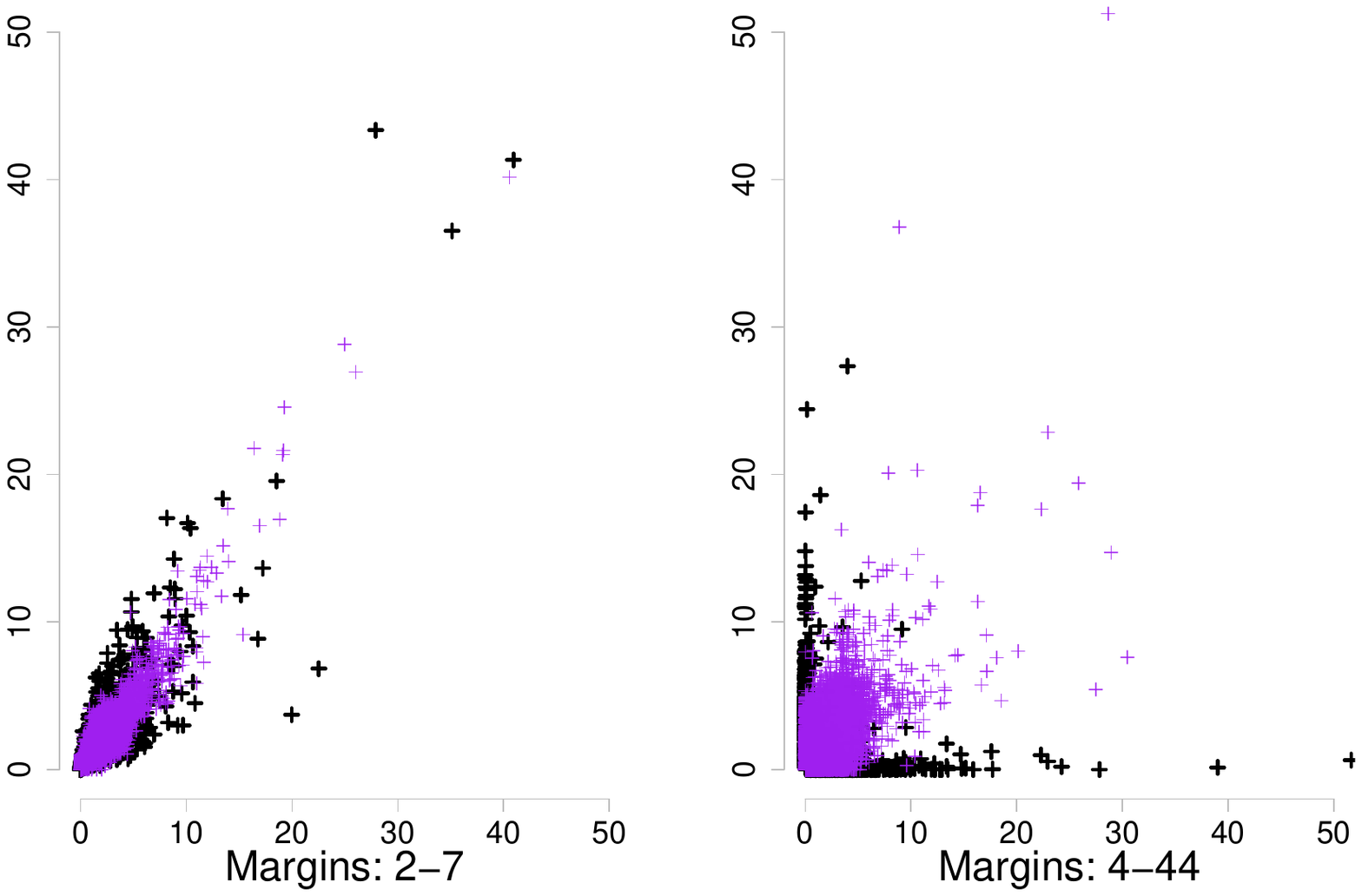}
        \caption{Hüsler--Reiss dependence structure. Left: $d=10$; right: $d=50$. -- Test extremes are in black while generated extremes by WA-GAN are in purple. For each dimension, the left plot corresponds to a high tail dependence scenario while the right plot corresponds to a low tail dependence scenario.}
        \label{fig:extremes-HR}
\end{figure}

\subsubsection{Gaussian dependence structure}
\label{sec:simus:gaussian}

In this setup, the random vector $\X$ is generated  using a $d$-dimensional exchangeable Gaussian copula with correlation parameter $\rho \in [-1,1]$ and Pareto distributed marginals with parameter $\alpha > 0$. It is easy to verify that this data-generating process does not satisfy our assumptions as Assumption~\ref{ass:phi_interior} is not verified: all variables are asymptotically independent and the angular measure is concentrated on the vertices of the unit simplex. 

We consider various scenarios in which we assess the performance of WA-GAN. We take $d \in \{10,20,50\}$ and a Kendall's tau of $\tau \in \{1/4,1/2,3/4\}$, corresponding to the correlation parameters $\rho \in \{\sin(\pi\tau/2): \tau = 0.25,0.5,0.75\}$. For the margins, we consider $\alpha = 2$ in any scenario. The models are trained on $n_{\text{train}} = 10\,000$ observations and validated on $n_{\text{val}} = 5\,000$ observations. The performance assessment is done on $n_{\text{test}} = 20\,000$ points. We do not report other methods' performances, as our aim is mainly to see how we may adapt our methodology to be resilient against asymptotic independence in the data.

As our method relies on Aitchison coordinates for the open simplex, it is not able to produce new angles located exactly on the boundary of the simplex. However, one crucial observation is the following: lower tail dependence corresponds to heavier tails in the Aitchison coordinate space. This is illustrated in Figure~\ref{fig:alphas} with $d=3$ and $n = 1000$ data points sampled from a Dirichlet distribution with parameter vector $\alpha = (2,2,2)$ (top) 
and $\alpha = (0.1,0.1,0.1)$ (bottom).
In each situation, we display the raw simplex data, the corresponding coordinates in the orthonormal basis introduced in Proposition~\ref{prop:ONbaseSimplex}, and the frequency histogram of the first coordinate values.
\begin{figure}[h]
  \centering
  \begin{subfigure}[t]{0.32\textwidth}
    \includegraphics[width=\linewidth]{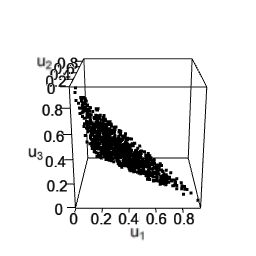}
    \caption{Raw simplex data}
    \label{fig:alpha2_simplex}
  \end{subfigure}
  \hfill
  \begin{subfigure}[t]{0.32\textwidth}
    \includegraphics[width=\linewidth]{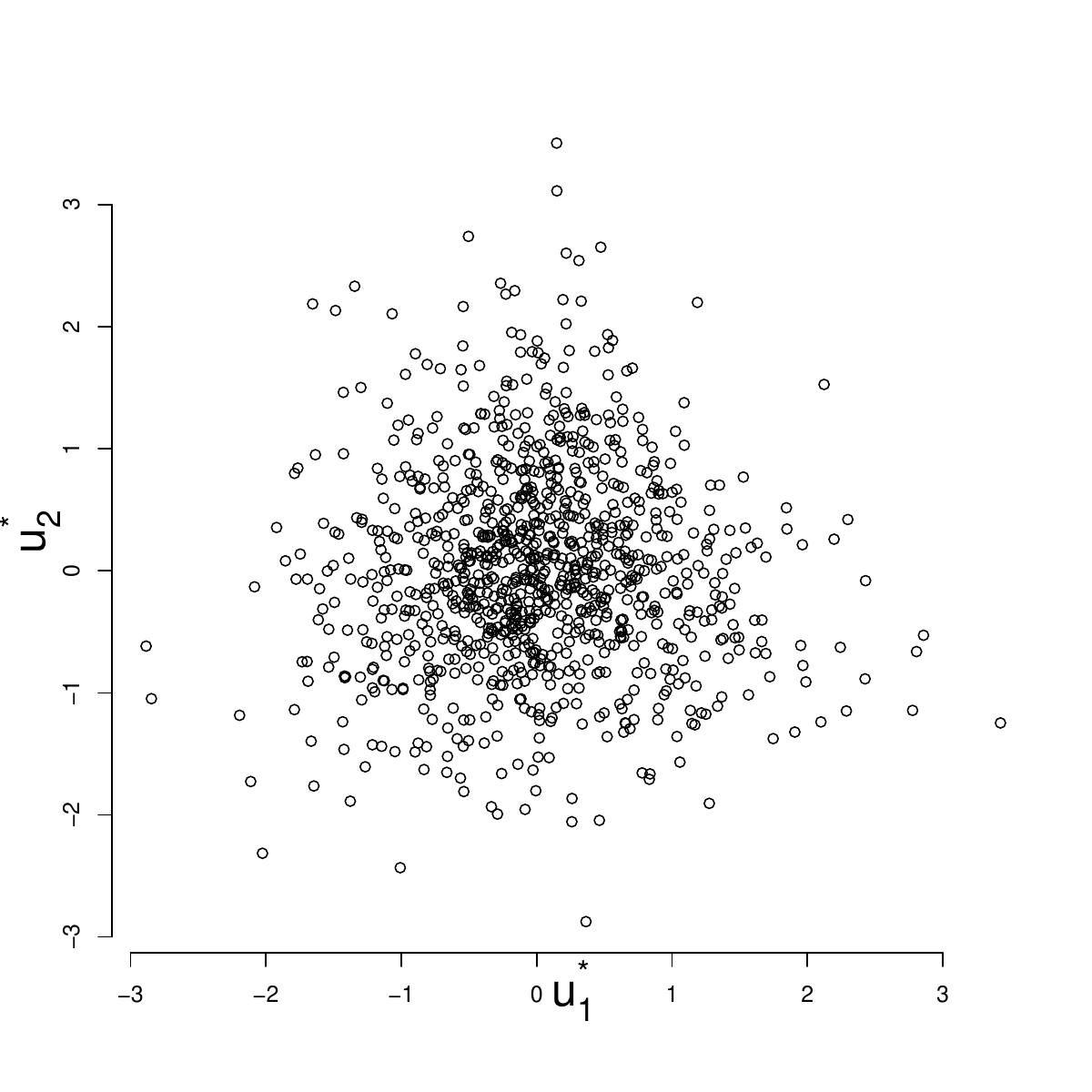}
    \caption{Aitchison coordinates}
    \label{fig:alpha2_coordinates}
  \end{subfigure}
  \hfill
  \begin{subfigure}[t]{0.32\textwidth}
    \includegraphics[width=\linewidth]{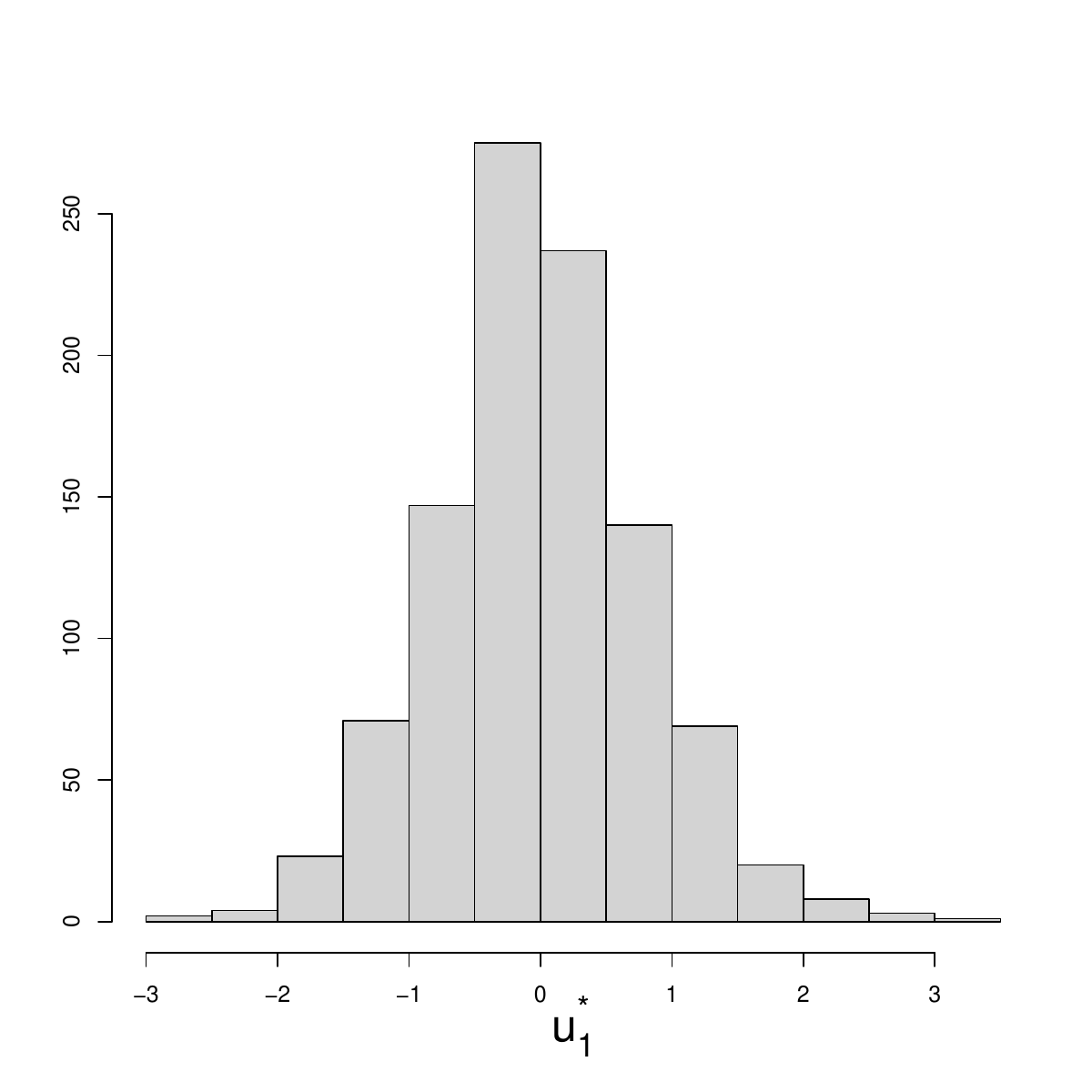}
    \caption{Histogram of the first coordinate values}
    \label{fig:alpha2_hist}
  \end{subfigure}
%
  \begin{subfigure}[t]{0.32\textwidth}
    \includegraphics[width=\linewidth]{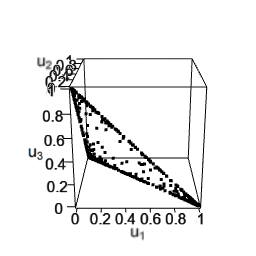}
    \caption{Raw simplex data}
    \label{fig:alpha01_simplex}
  \end{subfigure}
  \hfill
  \begin{subfigure}[t]{0.32\textwidth}
    \includegraphics[width=\linewidth]{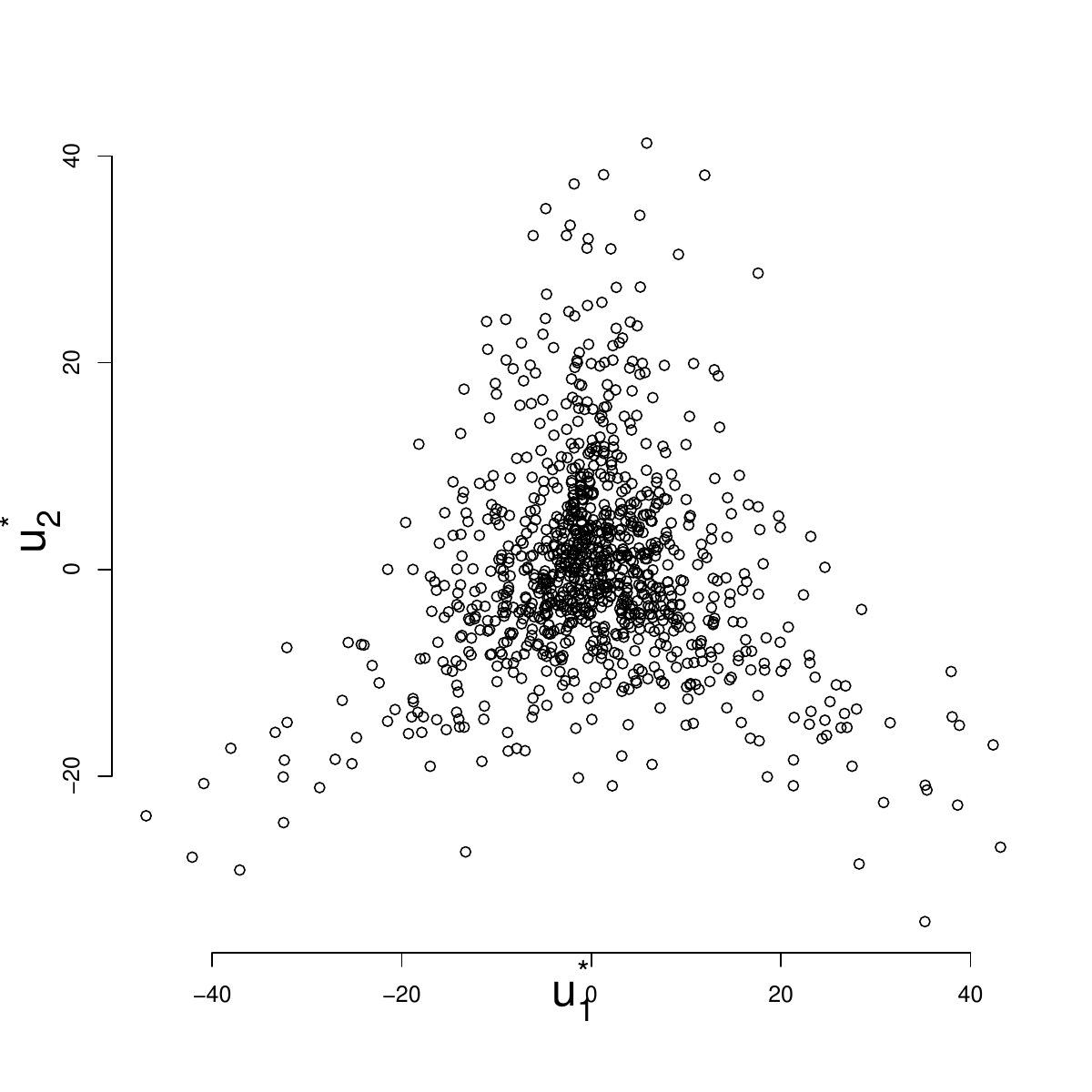}
    \caption{Aitchison coordinates}
    \label{fig:alpha01_coordinates}
  \end{subfigure}
  \hfill
  \begin{subfigure}[t]{0.32\textwidth}
    \includegraphics[width=\linewidth]{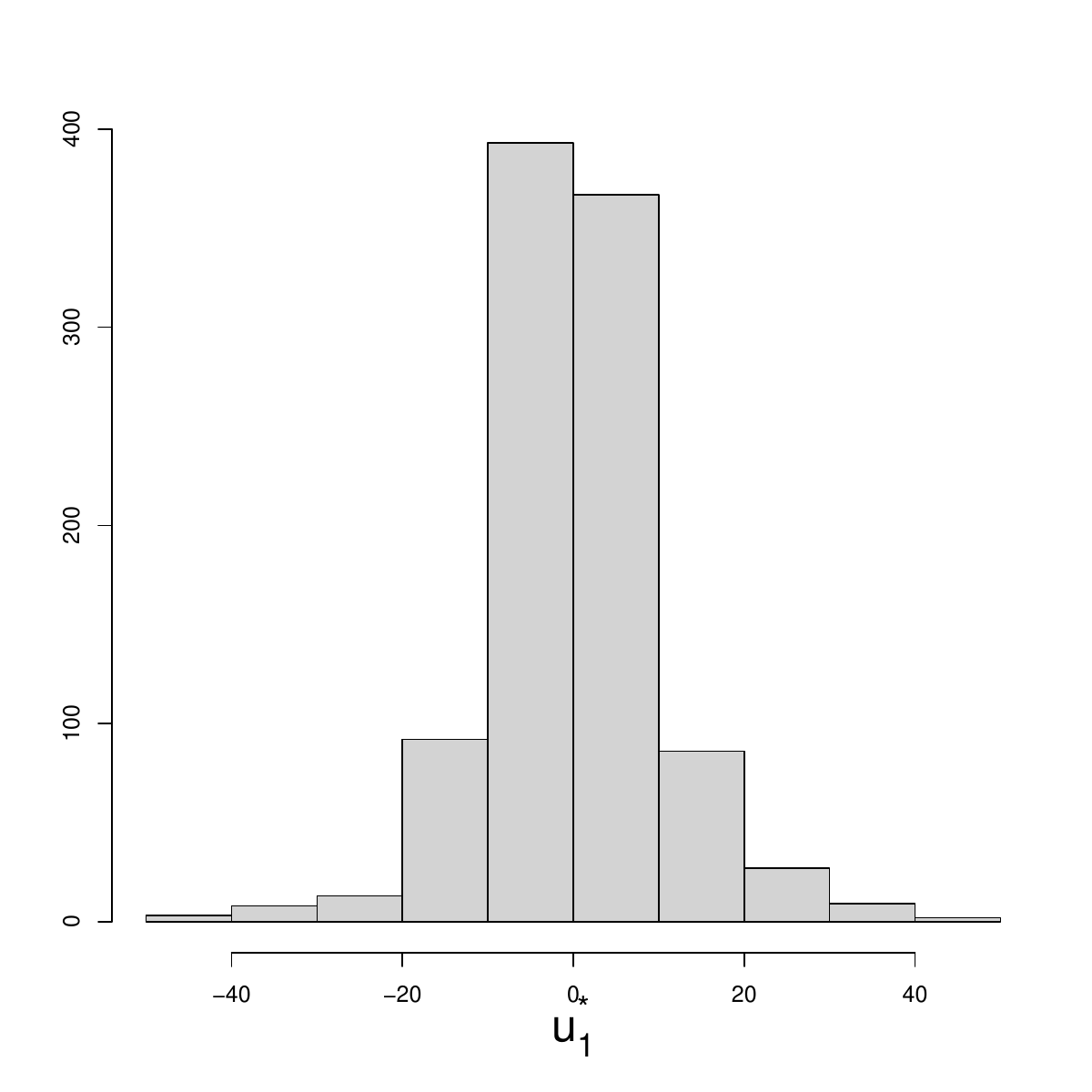}
    \caption{Histogram of the first coordinate values}
    \label{fig:alpha01_hist}
  \end{subfigure}

  \caption{The Aitchison transform for Dirichlet distributed data in $d=3$ with parameter vector $\alpha = (2,2,2)$ (top row) $\alpha = (0.1,0.1,0.1)$ (bottom row).}
  \label{fig:alphas}
\end{figure}

Given this observation, we propose to 
allow for other latent distributions than the standard normal one as input to the generator of WA-GAN. Here, we consider the Student distribution with possible degrees of freedom equal to $1$, $2.5$, $5$, or $\infty$, the latter value corresponding to the Gaussian distribution as before. 
The latent distribution choice is considered as a hyperparameter and is chosen randomly during the random search. 

\paragraph{Results.}

Table~\ref{tab:gan_scores_gaussian} reports the test scores, for the same two scores as used previously, of WA-GAN in each of the considered scenarios.

\begin{table}[h]
\centering
\begin{tabular}{l|ccc|ccc}
\toprule
 & \multicolumn{3}{c|}{Dependence score} & \multicolumn{3}{c}{Extremes score} \\
$\tau$ & $d=10$ & $d=20$ & $d=50$ & $d=10$ & $d=20$ & $d=50$ \\
\midrule
\multirow{1}{*}{$\tau = \frac{1}{4}$}  
& 0.0185 & 0.0214 & 0.0259 & 36.939 & \phantom{0}63.489 & 45.041 \\
\multirow{1}{*}{$\tau = \frac{1}{2}$} 
& 0.0335 & 0.0173 & 0.0166 & 53.859 & 117.450 & 71.547 \\
\multirow{1}{*}{$\tau = \frac{3}{4}$} 
& 0.0135 & 0.0115 & 0.0108 & 39.144 & \phantom{0}99.651 & 74.948 \\
\bottomrule
\end{tabular}
\caption{Gaussian dependence structure. Dependence and extremes scores for different dependence $\tau$ and dimensions $d$.}
\label{tab:gan_scores_gaussian}
\end{table}

The dependence score values are quite similar to those obtained with the logistic model. However, regarding extreme values, the results are worse, as the method tends to produce overly dependent tail observations (Figure~\ref{fig:d=10gaussian}). One possible explanation for the fact that the dependence score is not strongly affected is that, under asymptotic independence in the tail, the coefficients are very poorly estimated—even in the test set—so that the score values are not truly reliable in this case (Figure~\ref{fig:d=10gaussian_coefs}).
\begin{figure}[h]
    \centering
    \begin{subfigure}[b]{0.49\linewidth}
        \includegraphics[width=\linewidth]{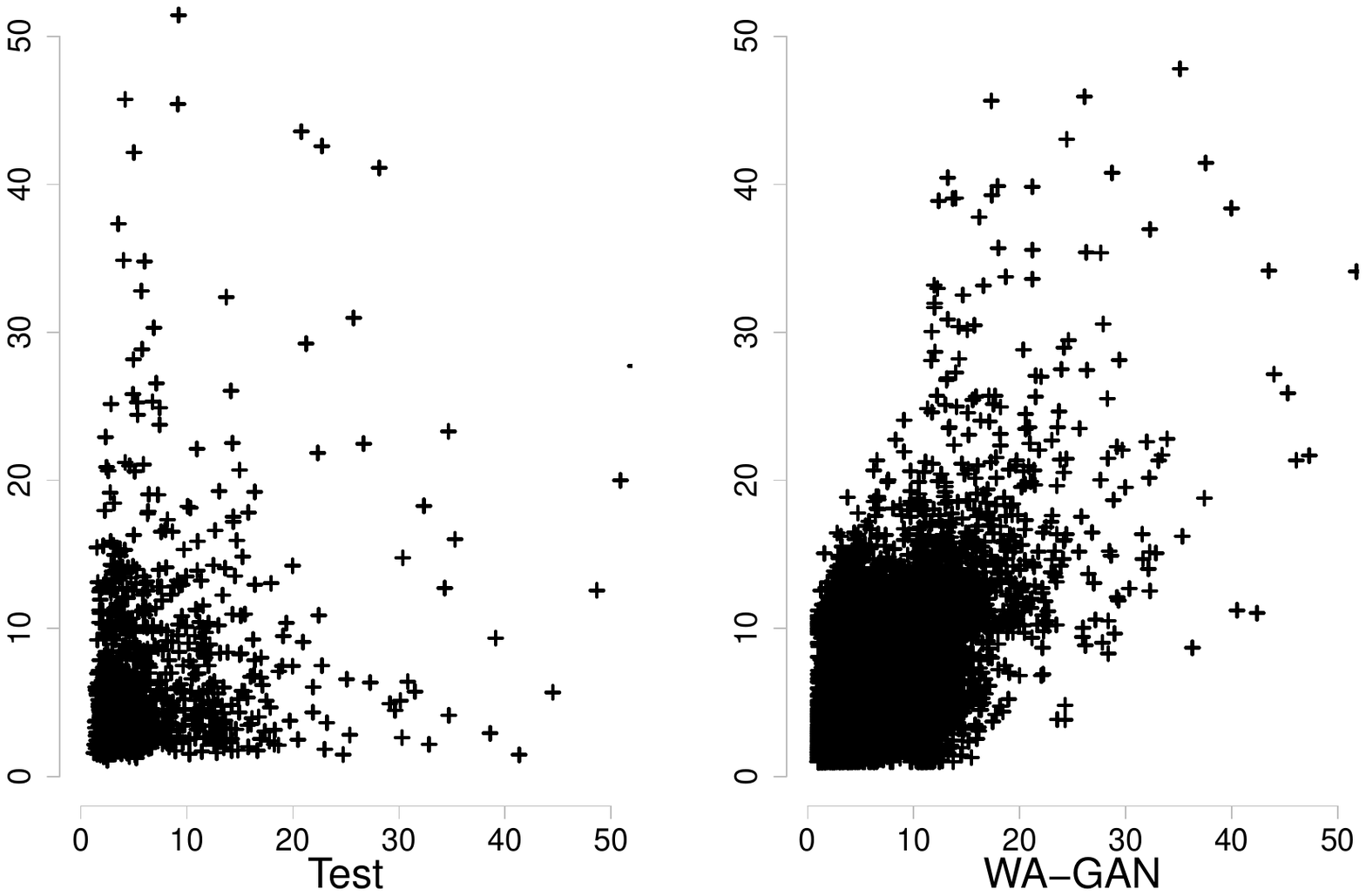}
        \caption{Gaussian dependence structure. First two margins of extremes in the test set and the generated ones using WA-GAN for $d=10$.}
        \label{fig:d=10gaussian}
    \end{subfigure}
    \hfill
    \begin{subfigure}[b]{0.49\linewidth}
        \includegraphics[width=\linewidth]{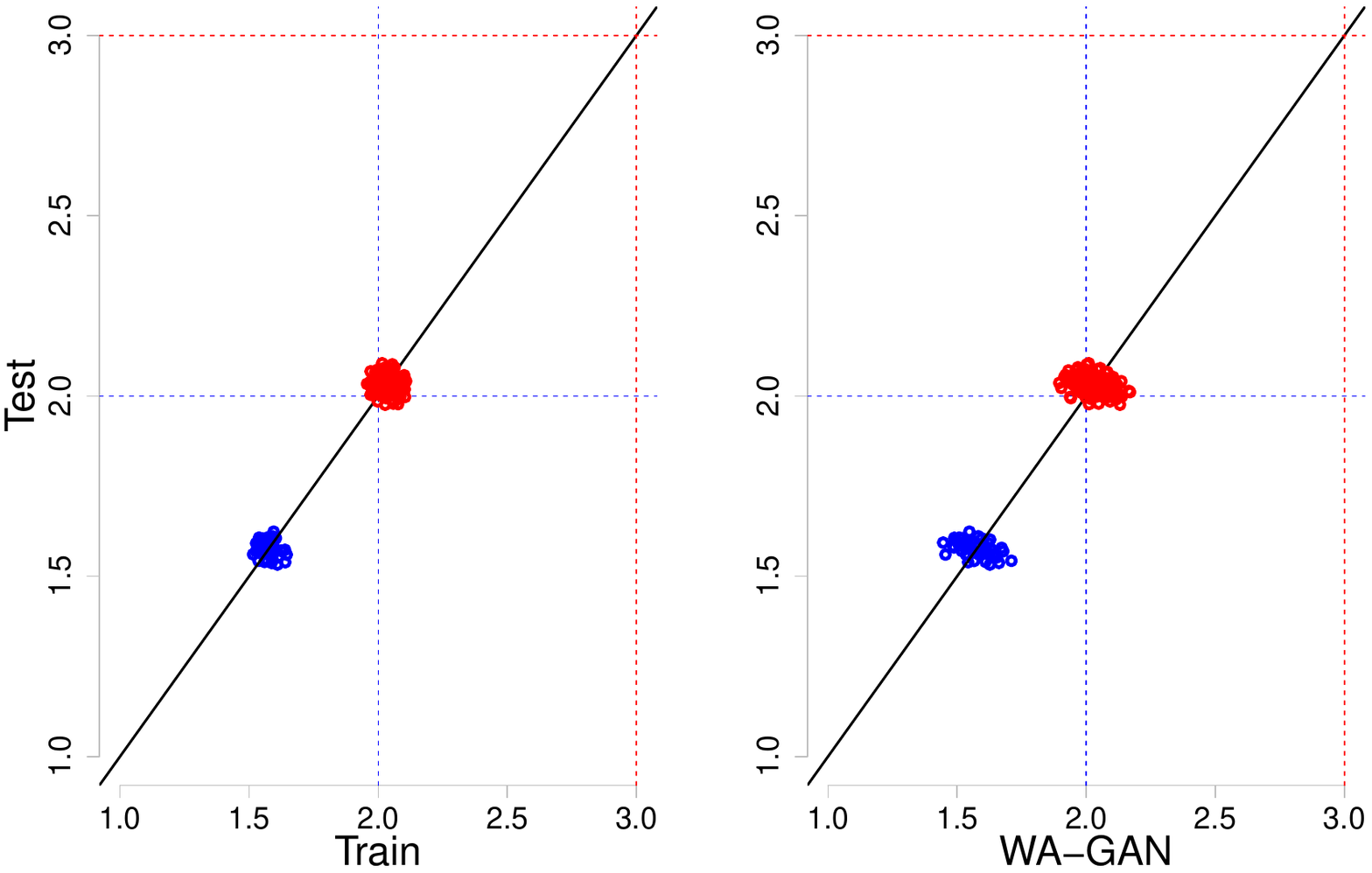}
        \caption{Gaussian dependence structure. Extremal coefficients of order $k=2$ (blue) and $k=3$ (red). Solid black line is the diagonal. Dashed blue and red lines are the true values of the coefficients for $k=2$ and $k=3$ respectively.}
        \label{fig:d=10gaussian_coefs}
    \end{subfigure}
    \caption{Results for Gaussian dependence structure with $d=10$.}
\end{figure}

Despite the rather poor performance in generating asymptotically independent extremes, it is remarkable that, out of the nine possible scenarios, six of the validated architectures employed a Student rather than Gaussian latent distribution for the Aitchison coordinates.
This suggests that allowing for heavier-tailed distributions may be beneficial when asymptotic independence is suspected. The selected degrees of freedom were four times $2.5$ and two times $5$, indicating that an excessively heavy-tailed distribution, with no finite mean, might be somewhat too extreme.

\egroup

\subsection{Financial data: daily returns of industry portfolios}
\label{sec:real}

We apply our methodology to analyze the tail behavior of the ``value-averaged'' daily returns of $d=30$ industry portfolios compiled and posted as part of the Kenneth French Data Library. The data in consideration span between 1950 and 2015 with $n = 16\,694$ observations. This data set is very widely used; for analyses in an extreme value context see e.g.\ \cite{janssen2020} and \cite{cooley2019}. Details about the portfolio constructions can be found online.\footnote{\url{https://mba.tuck.dartmouth.edu/pages/faculty/ken.french/Data_Library/det_30_ind_port.html}} Since we are interested in extreme losses we first multiply all returns by $-1$.

\begin{figure}[h]
    \centering
    \includegraphics[width=0.7\linewidth]{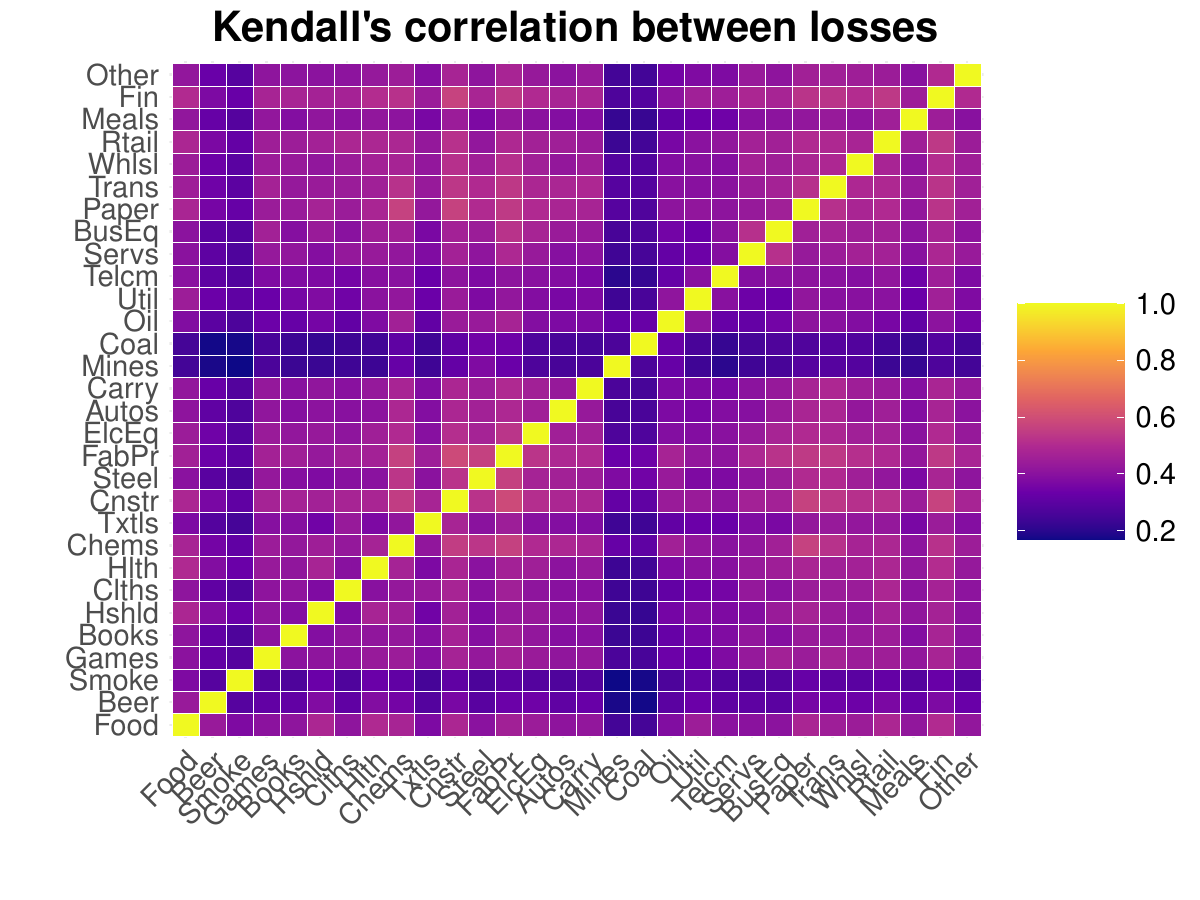}
    \caption{Kendall's correlation matrix between daily losses of the portfolios.}
    \label{fig:cor_matrix}
\end{figure}

As illustrated by the Kendall's correlation matrix in Figure~\ref{fig:cor_matrix}, the daily losses are positively related between the portfolios. We aim to investigate this dependence in the tail part of the distribution.

To sample extremes from the joint distribution, we first compute the maximum likelihood estimators of the univariate GP parameters for each margin. Boxplots illustrating the values of those parameters are reported in Figure~2 in Appendix~E. The estimated shape values clearly indicates the presence of heavy tails in the marginal distributions. The GP qq-plots of the marginal fits are to be found on Figures~3, 4 and~5 in Appendix~E in the supplement. They indicate a relatively good fit, even though some extreme quantiles are larger than the ones associated with the GP model.

For comparison, we start by computing the test scores for  WA-GAN, HTGAN and GPGAN. The training and validation procedures are exactly the same as described in Section~\ref{sec:simus}. Here, we train on $n_{\text{train}} = 7\,000$, we validate on $n_{\text{val}} = 3\,000$ and compute the scores on $n_{\text{test}} = 6\,694$ observations. We again consider $k = \sqrt{n_{\text{train}}}$ for WA-GAN. For HTGAN, we do not compute the scores on extremes because the marginal distributions are not known, see Appendix~D. Table~\ref{tab:scores_real} shows that WA-GAN seems to be the best at capturing the dependence between extremes in the data while GPGAN has a better $2$-Wasserstein distance score.
\begin{table}[h]
    \centering
        \begin{tabular}{c|c@{\qquad}c}
            \toprule
            method & dependence & extremes \\ \midrule
            WA-GAN & $\bm{0.018}$ & 11.93  \\ 
            HTGAN & 0.026 &  / \\ 
            GPGAN & 0.043 &  $\bm{8.46}$ \\ \bottomrule
        \end{tabular}
    \caption{Test scores of each method on the $d=30$ financial dataset. The best score (i.e., the lowest) is indicated in bold.}
    \label{tab:scores_real}
\end{table}

\begin{figure}[H]
    \centering
    \begin{subfigure}[b]{0.49\linewidth}
        \includegraphics[width=\linewidth]{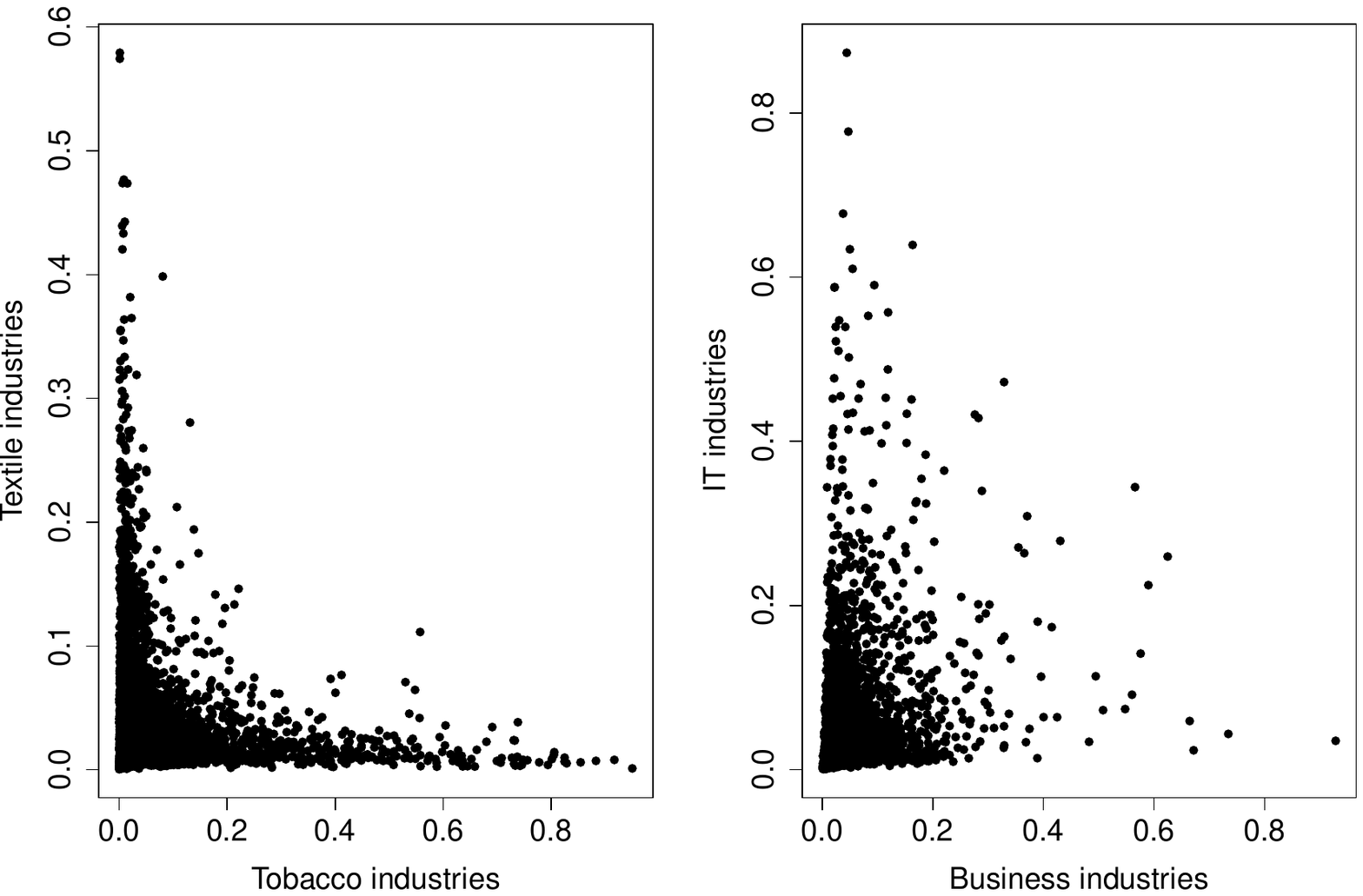}
        \caption{Two-dimensional projections of the generated ``angles'' from $\Phi$ by WA-GAN.}
        \label{fig:dep_finance}
    \end{subfigure}
    \hfill
    \begin{subfigure}[b]{0.49\linewidth}
        \includegraphics[width=\linewidth]{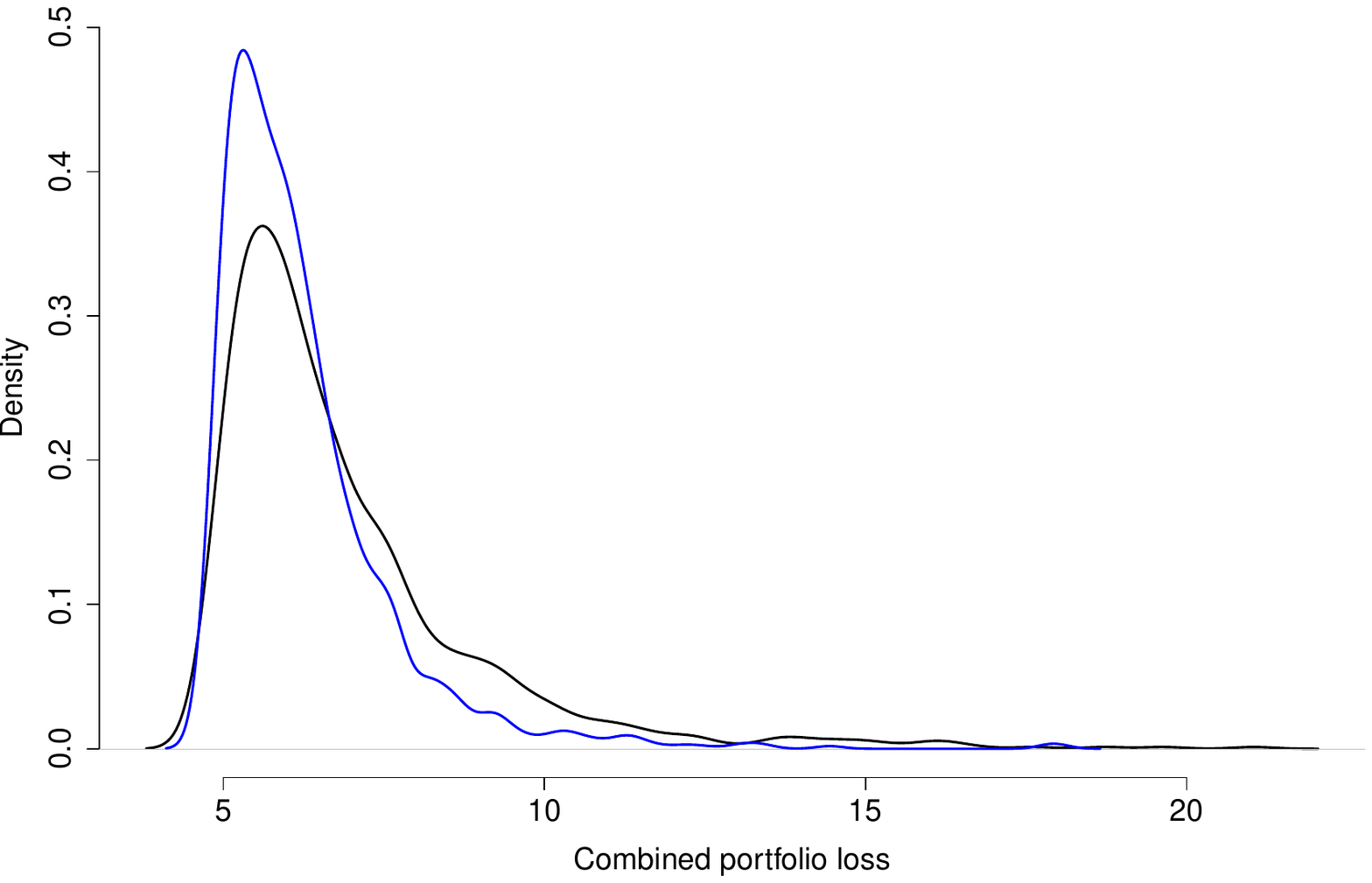}
        \caption{Densities of extreme losses in the combined portfolio of mines and coals industries based on WA-GAN (black) and independently simulated margins (blue).}
        \label{fig:combined_loss}
    \end{subfigure}
    \caption{Simulation results for the financial dataset.}
\end{figure}

As already pointed out in the analysis of~\cite{janssen2020}, the extreme losses of those portfolios can be clustered into different kind of industries. For example, the tobacco industry exhibits asymptotic independence to all other categories (e.g., to the textile industries) while the extreme losses of the business and IT-related industries are dependent. We illustrate the fact that WA-GAN also captures those phenomena in Figure~\ref{fig:dep_finance} by displaying the two-dimensional projections of the estimated angular measure on those margins. We see that the generated ``angles'' associated with tobacco/textile portfolios are much more concentrated around the axes than for business/IT, showing that WA-GAN is able to capture quite reasonably this complex tail dependence in this setting.

As a further illustration we consider a portfolio obtained by combining the mines industry portfolio with the coal industry portfolio (with equal weight). It is intuitive and has been shown in~\cite{cooley2019, janssen2020} that the mines and coal industries exhibit asymptotic dependence. If one wants to sample the losses associated to the portfolio we propose, this tail dependence should be taken into account. Figure~\ref{fig:combined_loss} represents the densities of simulated values of the extreme losses of the combined portfolio (i.e., both marginal losses exceeds a large threshold) based on WA-GAN (in black) and based on independent simulations from the marginal GP distributions (in blue). Not taking the tail dependence into account leads to underestimation of the risk associated with this portfolio.

\section{Conclusions}
\label{sec:conclusions}

This paper develops the WA-GAN method which is built on the $L_1$-norm and on the asymptotic independence of the angular and radial parts of a $d$-dimensional regularly varying random vector. An important component is the use of Aitchison coordinates which transforms values on the unit simplex in $\R^d$ to the entire linear space $\R^{d-1}$. The method provides simulated extreme values and can hence be used to estimate probabilities of extreme events, even more extreme than those already observed.

\bgroup
\color{changecol}
The method is tested on simulated extreme-value datasets of dimensions $d = 10$, $20$, and $50$, with $10{,}000$ observations in the training set and $5{,}000$ in the validation set. Its performance is evaluated using two metrics: a dependence score based on extremal coefficients and a Wasserstein score illustrating the quality of the generated extremes. In both cases, WA-GAN performs reasonably well compared to other existing methods such as HTGAN and GPGAN. However, its performance may deteriorate under asymptotic independence, as it relies on the Aitchison space, where low tail dependence corresponds to heavier-tailed coordinates, while WA-GAN uses a Gaussian latent distribution. Possible improvements to address this limitation will be discussed in future work.
\egroup

The methods are also applied to a financial data set from the Kenneth French Data Library. Again WA-GAN performs well and underlines the importance of being able to handle dependence between different financial portfolios. In terms of performance measures, WA-GAN has the best dependence score, while GPGAN has a better 2-Wasserstein score than WA-GAN.

In this paper we do not quantify the uncertainty in the estimated probability of an extreme event. However, provided enough computational power is available, this could be done using a approach similar to a parametric bootstrap. For this one would repeat the following, say, 100 times:  first use the fitted WA-GAN to simulate a new extreme data set of the same size as the original one, then fit a new WA-GAN to this simulated data set and use the new WA-GAN to estimate the probability of interest. This will provide 100 estimated probability values which can be used to find ``confidence bounds'' for the original estimated probability.

\bgroup
\color{changecol}
An alternative approach would be to take a Bayesian perspective: choose a prior distribution for the marginal tail parameters and the angular measure (or any other means of describing the dependence structure) and then simulate from the predictive distribution given the data: first simulate random tail and parameters from the posterior distribution and then sample from the resulting multivariate generalized Pareto distribution. The advantage of such an approach is that the estimation uncertainty would be taken into account through the posterior distribution of the tail and dependence parameters. Drawbacks are that the construction of a suitable prior on the dependence structure may become quite delicate and the generation from the posterior distribution rather involved, especially in high dimensions.

Another important aspect, not addressed in this work but potentially relevant in applications, is the possible time-evolving nature of the extremal dependence structure of the data \cite{deCarvalho2025techs, drees2023time}, or its variation with respect to a latent variable. Covariates can be incorporated into (Wasserstein) GANs by using their conditional versions \cite{mirza2014cGAN, zheng2020cWGANGP}, which could facilitate the generation of new multivariate extremes for specific covariate values, like time, in a manner similar to what has been explored for image generation in \cite{gauthiercGAN}. These directions are left for future work.
\egroup

\section*{Code and data availability}

The code is publicly available on the repository \url{https://github.com/stephanelh98/extreme-WGAN}. The data underlying the analysis in Section~\ref{sec:real} is publicly available from the Kenneth R. French data library \url{https://mba.tuck.dartmouth.edu/pages/faculty/ken.french/data_library.html}.

\section*{Acknowledgments}

The authors sincerely thank the anonymous reviewers for their insightful comments and suggestions, which have significantly improved the quality and clarity of this paper.

The research of St{\'e}phane Lhaut was supported by the Fonds National de la Recherche Scientifique (FNRS, Belgium) within the framework of a FRIA grant (No. 1.E.114.23F).
Computational resources have been provided by the Consortium des Équipements de Calcul Intensif (CÉCI), funded by the Fonds de la Recherche Scientifique de Belgique (F.R.S.-FNRS) under Grant No.\ 2.5020.11 and by the Walloon Region.

\bibliographystyle{plain}
\bibliography{ref}

\appendix

\section{Proof of Proposition~\ref{prop:MGPapprox}}
\label{app:proof_prop1}

We start by noting that $\X \eqlaw \bm{b}(\V)$ and, by the continuity of the margins $F_j$, each tail quantile function $b_j$ is strictly increasing, so that $\{\bm{\bm{X}} \not\leq \bm{b}(t)\} = \{\V \nleq t\}$ for any $t>1$. Consequently,
\[
    \frac{\X - \bm{b}(t)}{\bm{a}(t)} \, \Big| \, \X \not\leq \bm{b}(t)
    \eqlaw \frac{\bm{b}(t(t^{-1}\V))-\bm{b}(t)}{\bm{a}(t)} \, \Big| \, \V \nleq t.
\]
Furthermore, it follows easily from Assumption~\ref{ass:multRV} that 
\begin{equation}
\label{eq:tVVt}
    \lp t^{-1} \V \mid \V \not\le t \rp
    \wc \Y, \qquad t \to \infty,
\end{equation}
where, for Borel set $B \subseteq \EE$, writing $\LL_1 \de \{ \x \in [0, \infty)^d : \x \not\leq 1 \}$, we have
\begin{equation}
\label{eq:Ynu}
    \P(\Y \in B) = \frac{\nu(B \cap \LL_1)}{\nu(\LL_1)}.
\end{equation}
The exponent measure $\nu$ is connected to the angular measure $\Phi$ introduced in~\eqref{eq:RVpolar}. More precisely~\cite[Chapter~8]{beirlant2004}, for any bounded, measurable function $f : \EE \to \R$ that is zero in a neighborhood of the origin, we have
\begin{equation}
\label{eq:polar_integral}
    \nu(f) 
    \de \int_\EE f(x) \, \diff \nu(x)
    = d \int_{\simplex} \int_0^\infty f(y\w) \frac{\diff y}{y^2} \, \diff \Phi (\w).
\end{equation}
In particular, for any for Borel set $B \subseteq \EE$,
\begin{align*}
    d^{-1} \nu(B \cap \LL_1)
    &=
    \int_{\simplex} \int_0^\infty \I_{B \cap \LL_1}(y\w) \frac{\diff y}{y^{2}} \, \diff \Phi(\w) \\
    &=
    \int_{\simplex} \int_1^\infty \I_{B \cap \LL_1}(y\w) \frac{\diff y}{y^2} \, \diff \Phi(\w) \\
    &= \P(Y \bTheta \in B \cap \LL_1),
\end{align*}
where $Y$ is unit-Pareto distributed and independent of $\bTheta \sim \Phi$. Hence, by \eqref{eq:Ynu},
\[
    \P(\Y \in B)
    = \frac{d \, \P(Y \bTheta \in B \cap \LL_1)}{d \, \P(Y \bTheta \in \LL_1)}
    = \P\lp Y \bTheta \in B \mid Y \bTheta \in \LL_1 \rp,
\]
From the local uniformity of the convergence in point~(ii) of Remark~\ref{rem:equiv_margins}, guaranteed by Assumption~\ref{ass:margins_rv}, and the fact that $\P(Y_j > 0) = 1$ for any $j \in \{1,\ldots,d\}$ by the previous computations and Assumption~\ref{ass:phi_interior}, the extended continuous mapping theorem~\cite[Theorem~1.11.1]{van1996weak} applies, yielding
\[
    \frac{\bm{b}(t(t^{-1}\V))-\bm{b}(t)}{\bm{a}(t)} \, \Big| \, \V \nleq t
    \wc \frac{\bm{Y}^{\bm{\xi}}-1}{\bm{\xi}},
    \qquad t \to \infty.
\]
This concludes the proof of~\eqref{eq:XbtatYxi} and~\eqref{eq:YdYW}. Eq.~\eqref{eq:wcYu} is a direct consequence of~\eqref{eq:XbtatYxi} as
\[
    \sigma_j(u_j(t)) 
    = a_j \lp \frac{1}{1-F_j(b_j(t))} \rp 
    = a_j(t), \qquad t > 1
\]
since $F_j(F_j^{-1}(p)) = p$ for all $0 < p < 1$ by continuity of $F_j$. \qed

\section{The Aitchison simplex}
\label{app:aitchison}

The open simplex $\osimplex$ is equipped with an inner product space structure with the following operations~\cite{aitchison1982}: for $\v = (v_1,\ldots,v_d) , \w = (w_1,\ldots,w_d) \in \osimplex$ and $\alpha \in \R$, define the following operations:
\begin{align*}
    \v \oplus \w 
    &\de \lp \frac{v_j w_j}{\sum_{i=1}^d v_i w_i} \rp_{j=1}^d
    && \text{(addition),}\\
    \alpha \odot \v 
    &\de \lp \frac{v_j^\alpha}{\sum_{i=1}^d v_i^\alpha} \rp_{j=1}^d
    && \text{(scalar multiplication),} \\
    \left\langle \v, \w \right\rangle_A 
    & \de \sum_{i=1}^d \log \lp \frac{v_i}{g(\v)} \rp \log \lp \frac{w_i}{g(\w)} \rp
    && \text{(Aitchison inner product),}
\end{align*}    
where $g : \u \in \osimplex \mapsto  g(\u) \de (\prod_{i=1}^d u_i)^{1/d}$ is the geometric mean. 
It is an easy exercise to verify that $(\osimplex, \oplus, \odot)$ forms a real vector space with zero element $\0_A = \1_d/d \in \osimplex$, where $\1_d = (1,\ldots,1) \in \R^d$, and that $\left\langle \cdot, \cdot \right\rangle_A$ defines an inner product on it. The open simplex equipped with this structure is often called the \emph{Aitchison simplex}. 

A trivial but useful observation is that if one introduces the so-called \emph{Centered LogRatio (CLR)} operator
\begin{equation}
\label{eq:clr}
    \clr : \u \in \osimplex \mapsto \clr(\u) \de \lp \log \lp \frac{u_j}{g(\u)} \rp \rp_{j=1}^d \in \H \subset \R^d,
\end{equation}
where $\H \de \{\x \in \R^d : \langle \x, \1_d\rangle = 0\}$ with $\langle \cdot,\cdot\rangle$ the standard inner product in $\R^d$, then
\[
    \left\langle \v, \w \right\rangle_A = \langle \clr(\v), \clr(\w) \rangle, \qquad \v,\w \in \simplex,
\]
so that $\clr$ naturally defines an isometry between $(\osimplex, \langle \cdot,\cdot\rangle_A)$ and $(\H, \langle \cdot,\cdot \rangle)$. The inverse transformation $\clr^{-1} : \H \mapsto \osimplex$ is the well known \emph{softmax} function
\[
    \clr^{-1}(\x) =  \lp \frac{\exp(\x_j)}{\sum_{i=1}^d \exp(\x_i)} \rp_{j=1}^d = \operatorname{softmax}(\x), \qquad \x \in \H.
\]

The above considerations lead to a clear and easy way to construct an orthonormal basis of the Aitchison simplex:
\begin{enumerate}
    \item Take any linearly independent family $\{\x_1, \ldots, \x_{d-1}\}$ in $\H$.
    \item Apply the Gram--Schmidt procedure to produce an orthonormal basis $\{\e_1, \ldots, \e_{d-1}\}$ of $\H$ with respect to the standard inner product on $\R^d$.
    \item Compute $\{\e_i^* \de \clr^{-1}(\e_i) : i = 1,\ldots,d-1\}$, which forms an orthornormal basis of the Aitchison simplex.
\end{enumerate}
Applying the above procedure to the free family
\[
    \bigl\{ \x_i \de (0,\ldots,0,1,-1,0,\ldots,0 ) \in \H : i = 1,\ldots, d-1 \bigr\},
\]
where in $\x_i$ the $1$ element lies at the $i$-th position, leads to the orthonormal basis of the Aitchison simplex presented in Proposition~\ref{prop:ONbaseSimplex}.

\section{Angular measure with respect to another norm}
\label{app:change_norms}

The angular measure $\Phi$ in~\eqref{eq:RVpolar} can be introduced with respect to an arbitrary norm $\|\cdot\|$ on $\R^d$, that is, if we let
\[
    \widetilde{R} = \|\V\|  \qquad \text{and} \qquad \widetilde{\W} = \widetilde{R}^{-1} \V,
\]
one can consider the measure $\tPhi$ on $\widetilde{\Delta}_{d-1} \de \{\x \in [0,1]^d : \|\x\| = 1\}$ obtained by taking, as in~\eqref{eq:RVpolar},
\begin{equation}
\label{eq:rvPolarTilde}
    \tPhi(A) = \lim_{t \to \infty} \P \lp \widetilde{\W} \in A \mid \widetilde{R} \geq t\rp,    
\end{equation}
for any Borel set $A \subseteq \widetilde{\Delta}_{d-1}$ such that $\tPhi(\partial A) = 0$. Popular choices for $\|\cdot\|$ consist of the $L_p$ norms $|\cdot|_p$ on $\R^d$ for $p \in [1,\infty]$, see, e.g.,~\cite{einmahl2001nonparametric,einmahl2009maximum,clemencon2023concentration}. Even though we only consider the norm $|\cdot|_1$ in Algorithm~\ref{alg:WGAN}, we show that a simple post-processing step permits to provide an estimate for $\tPhi$ also. 

From~\eqref{eq:rvPolarTilde}, one may show that $\tPhi(A) = \Psi(A)/\Psi(\widetilde{\Delta}_{d-1})$ where $\Psi$ is the finite Borel measure on $\widetilde{\Delta}_{d-1}$ obtained by taking
\[
    \Psi(A) = \lim_{t \to \infty} t \ \P \lp \widetilde{\W} \in A, \widetilde{R} \geq t\rp
\]
provided $\Psi(\partial A) = 0$.
From~\eqref{eq:polar_integral}, for any bounded, measurable $f : \EE \to \R$ vanishing in a neighborhood of $\0$, we get
\[
    d \int_{\simplex} \int_0^\infty f(r\w) \frac{\diff r}{r^2} \, \diff\Phi(\w)
    = \int_{\widetilde{\Delta}_{d-1}} \int_0^\infty f(r\widetilde{\w}) \frac{\diff r}{r^2} \, \diff \Psi(\widetilde{\w}),
\]
Taking $f(x) = g(x/\|x\|) \I(\|x\| \geq1)$ for $x \in \EE$, where $g : \widetilde{\Delta}_{d-1} \to \R$ is some bounded function, we get
\[
    d \int_{\simplex} g \lp \frac{\w}{\|\w\|} \rp \| \w \| \diff\Phi(\w) 
    = \int_{\widetilde{\Delta}_{d-1}} g(\widetilde{\w})  \diff \Psi(\widetilde{\w}).
\]
Picking $g(\widetilde{\w}) = \I(\widetilde{\w} \in \widetilde{\Delta}_{d-1})$ gives
\[
    \Psi(\widetilde{\Delta}_{d-1}) = d \int_{\simplex} \| \w \| \, \diff\Phi(\w),
\]
so that for any bounded $g : \widetilde{\Delta}_{d-1} \to \R$ we have
\begin{equation}
\label{eq:tPhilink}
    \int_{\widetilde{\Delta}_{d-1}} g(\widetilde{\w}) \, \diff \tPhi(\widetilde{\w})
    = \frac{\int_{\simplex} g \lp \frac{\w}{\|\w\|} \rp \| \w \| \, \diff\Phi(\w)}{ \int_{\simplex} \| \w \| \, \diff\Phi(\w)}.
\end{equation}
In particular, $\tPhi$ is fully determined by $\Phi$. 

Suppose that we observe $\bTheta_1,\ldots,\bTheta_n \sim \Phi$ as it would be (approximately) the case via the trained generator in Algorithm~\ref{alg:WGAN}. They lead to the empirical estimator $\hPhi = (1/n) \sum_{i=1}^n \delta_{\bTheta_i}$ for $\Phi$. Relation~\eqref{eq:tPhilink} then justifies the approximation
\begin{align*}
    \int_{\widetilde{\Delta}_{d-1}} g(\widetilde{\w}) \, \diff \tPhi(\widetilde{\w}) 
    &\approx \frac{\int_{\simplex} g \lp \frac{\w}{\|\w\|} \rp \| \w \| \, \diff\hPhi(\w)}{ \int_{\simplex} \| \w \| \, \diff\hPhi(\w)} \\
    &= \sum_{i=1}^n \Lambda_i g \lp \frac{\bTheta_i}{\|\bTheta_i\|} \rp 
    = \int_{\widetilde{\Delta}_{d-1}} g(\widetilde{\w}) \, \diff \widehat{\tPhi}(\widetilde{\w}),
\end{align*}
where
\[
    \widehat{\tPhi} \de \sum_{i=1}^n \Lambda_i \delta_{\bTheta_i / \|\bTheta_i\|}, \qquad \Lambda_i \de \frac{\|\bTheta_i\|}{\sum_{i=1}^n \|\bTheta_i\|}.
\]
Said otherwise, a sample $\bTheta_1,\ldots,\bTheta_n \sim \Phi$ can be used to produce an estimator for $\tPhi$ by considering the weighted empirical distribution of the rescaled angles $\bTheta_i$ on $\widetilde{\Delta}_{d-1}$, where the weights are proportional to the norm $\|\bTheta\|_i$. 

\end{document}